\documentclass[journal]{IEEEtran}
\ifCLASSINFOpdf
\else
\fi
\usepackage{graphicx}

\usepackage{threeparttable}
\usepackage{subfigure}

\usepackage{multicol}
\usepackage{multirow}
\newcommand{\mydeleted}[1]{\textcolor{red!30}{}}
\newcommand{\myadded}[1]{\textcolor{black}{#1}}
\usepackage{float}
\usepackage{CJK}
\usepackage[table]{xcolor}
\usepackage{amsmath,amssymb,amsfonts}
\usepackage{booktabs}
\usepackage{dsfont}
\newcounter{rownumbers}

\usepackage[ruled,linesnumbered]{algorithm2e}
\usepackage{verbatim}
\usepackage{color}
\usepackage{xcolor}    
\usepackage{wrapfig} 
\usepackage{cite}
\usepackage{enumitem}
\usepackage{url}
\usepackage{algorithmic}
\usepackage{dsfont}
\usepackage{pifont}
\usepackage{bm}
\usepackage[table]{xcolor}
\usepackage{colortbl}
\usepackage{graphicx}
\usepackage{subcaption}
\usepackage[colorlinks,linkcolor=red]{hyperref}
\begin{document}
%
\title{Bare Demo of IEEEtran.cls\\ for IEEE Journals}
%
%
%
\title{Unveiling the Backdoor Mechanism Hidden Behind Catastrophic Overfitting in Fast Adversarial Training}

\author{Mengnan~Zhao,
	Lihe~Zhang, Tianhang~Zheng, Bo~Wang, Baocai~Yin
\thanks{Manuscript received Sep, 2025.}
\thanks{This work was supported by the National Natural Science Foundation of China under Grant 62431004 and 62276046.}
\thanks{Mengnan Zhao is with the School of Computer Science and Technology, Anhui University, Hefei 230601, China. E-mail: gaoshanxingzhi@163.com.}
\thanks{Lihe Zhang and Bo Wang are with the School of Information and Communication Engineering, Dalian University of Technology (DUT), Dalian 116024, China. E-mail: zhanglihe@dlut.edu.cn, bowang@dlut.edu.cn.}
\thanks{Tianhang Zheng is with the School of Computer Science and Technology, Zhejiang University, Hangzhou 310058, China. E-mail: zthzheng@zju.edu.cn.}
\thanks{Baocai Yin is with the School of Computer Science and Technology, DUT, Dalian 116024, China. E-mail: ybc@dlut.edu.cn.}}
\markboth{IEEE Transactions on Information Forensics and Security,~Vol.~xx, No.~x, Month~20xx}%
{Zhao \MakeLowercase{\textit{et al.}}: Unveiling the Backdoor Mechanism Hidden Behind CO in FAT}

\maketitle

\begin{abstract}
Fast Adversarial Training (FAT) has attracted significant attention due to its efficiency in enhancing neural network robustness against adversarial attacks.   
However, FAT is prone to catastrophic overfitting (CO), wherein models overfit to the specific attack used during training and fail to generalize to others.
While existing methods introduce diverse hypotheses and propose various strategies to mitigate CO, a systematic and intuitive explanation of CO remains absent.
In this work, we innovatively interpret CO through the lens of backdoor.
Through validations on pathway division, diverse feature predictions, and universal class-distinguishable triggers in CO, we conceptualize CO as a weak-trigger variant of unlearnable tasks, unifying CO, backdoor attacks, and unlearnable tasks under a common theoretical framework.
Guided by this, we leverage several backdoor-inspired strategies to mitigate CO: (i) Recalibrate CO-affected model parameters using vanilla fine-tuning, linear probing, or reinitialization-based techniques; (ii) Introduce a weight outlier suppression constraint to regulate abnormal deviations in model weights. Extensive experiments support our interpretation of CO and show the efficacy of the proposed mitigation strategies.
\end{abstract}


%
\IEEEpeerreviewmaketitle

\section{Introduction}\label{sec:introduction}

\IEEEPARstart{R}{ecent} advancements in deep learning have driven significant progress across a wide range of applications \cite{zuo2022deep, mousavi2022deep, pereira2022sleap, baek2022deep, sapoval2022current}. However, these developments have also exposed critical limitations of neural networks \cite{chou2023backdoor, wei2022noise, chen2022effective}, particularly their susceptibility to adversarial attacks \cite{cao2022advdo, gu2022segpgd, zhong2022shadows}. To address such vulnerabilities, adversarial training has emerged as a widely adopted defense strategy \cite{yang2024structure,casper2024defending,zhang2024defensive,yue2024revisiting,fang2024enhancing,tang2024effective}, wherein perturbed examples are incorporated during training to enhance model robustness \cite{mo2022adversarial, jia2022adversarial, wu2022towards}.
Initial adversarial training techniques \cite{xiao2022stability,jin2022enhancing,guzman2022cross} produce training data using multi-step adversarial attacks, such as the projected gradient descent (PGD) \cite{madry2017towards}.
In recent years, fast adversarial training (FAT) methods \cite{pan2024adversarial,kim2021understanding}, particularly single-step approaches like FGSM-RS \cite{Wong2020}, offer notable computational efficiency by generating adversarial examples with fewer backward passes \cite{li2022subspace,huang2023fast,park2021reliably,yang2024fast}.

However, FAT remains susceptible to \textit{catastrophic overfitting (CO)}, where models overfit to the specific attack used during training and fail to generalize to other adversarial attacks. To address this, prior works primarily target surface-level symptoms—such as limited adversarial diversity \cite{Wong2020,huang2022fast}, disparities in sample convergence rates \cite{zhao2023fast}, and inconsistencies in gradient attribution importance \cite{golgooni2023zerograd}—to design corresponding mitigation strategies \cite{Andriushchenko2020,Jiag2022prior}. While these methods improve FAT stability, they offer limited insight into the underlying mechanisms of CO.
Beyond these, other studies have associated CO with deeper factors such as self-fitting behavior \cite{he2023investigating}, gradient misalignment \cite{andriushchenko2020understanding}, and feature overriding \cite{zhao2024catastrophic}. Collectively, these works suggest that CO stems from abnormal pathway division in the learned model representations. However, a systematic and intuitive explanation of CO remains absent, such as the mechanisms underlying pathway division and the nature of self-information.

This work innovatively analyzes CO through the lens of backdoor  \cite{zhang2024backdoor,yang2024watch,liang2024badclip,yin2024physical}.
We begin by validating the pathway division and diverse path predictions. These analyses reveal an intriguing phenomenon: CO exhibits a high similarity to backdoor-related tasks. We then show that adversarial perturbations from CO-affected models encode universal, class-discriminative triggers, and that the primary differences between CO and backdoor-related tasks can be attributed to the trigger strength variance. Together, these findings support a unified interpretation—trigger overfitting—in which both CO and backdoor-related tasks arise from a model’s over-reliance on trigger-like features transmitted through specialized paths.

We further introduce backdoor-inspired strategies to mitigate CO. (i) We adapt established backdoor fine-tuning techniques—including vanilla fine-tuning, linear probing, and reinitialization-based methods—to recalibrate the parameters of CO-affected models. These approaches steer the model away from overfitting by shifting its focus back to informative data features, rather than to trigger-related patterns. Experimental results show that fine-tuning  CO-affected models on clean data can temporarily alleviate CO, though the benefit is short-lived as CO tends to reoccur. (ii) Inspired by weight poisoning in backdoor attacks, we propose a weight outlier suppression constraint, which penalizes weight deviations from the layer-wise mean weight. 
This prevents the formation of adversarial pathways, leading to a stable FAT process.

Additionally, we provide a discussion section that (i) clarifies the essence of robustness improvement in adversarial training, (ii) analyzes the underlying cause of CO, and (iii) investigates whether techniques designed to mitigate CO can generalize to backdoor-related tasks.

In summary, our contributions are threefold:
\begin{itemize}[itemsep=0.2em, leftmargin=8pt]
\item[1)] 
We interpret CO through the lens of backdoor, unifying CO, backdoor attacks, and unlearnable tasks under a trigger overfitting framework.
\item[2)] We validate key similarities—pathway division, diverse predictions, and universal triggers—between CO and backdoor.
\item[3)] 
We introduce backdoor-inspired mitigation strategies, including adapted fine-tuning and a weight outlier suppression constraint, demonstrating their effectiveness empirically.
\end{itemize}

The remainder of the paper is organized as follows. Section \ref{related_work} reviews recent advances in adversarial training and backdoor-related tasks. In Section \ref{comparison}, we establish the connection between CO and backdoor-related tasks. Section \ref{sec:mitigation} presents a set of backdoor-inspired strategies for mitigating CO. Section \ref{DAA} offers further discussion and analysis, such as insights into the causes of CO and the robustness of FAT. Finally, Section \ref{conclusion} concludes the paper and outlines several promising directions for future research.

\section{Related work}\label{related_work}
\subsection{Adversarial training}
Recent breakthroughs in deep neural networks \cite{zuo2022deep,mousavi2022deep,pereira2022sleap} have prompted extensive research into their security risks \cite{chou2023backdoor,wei2022noise,chen2022effective}, with particular attention to their vulnerability to adversarial attacks \cite{Kurakin2017,dong2018boosting,cao2022advdo,gu2022segpgd,zhong2022shadows}. In response, adversarial training \cite{mo2022adversarial,jia2022adversarial,wu2022towards,kuang2024defense,wang2024revisiting} has emerged as a popular strategy to enhance model robustness, employing both multi-step (\textit{e.g.}, PGD \cite{madry2017towards}) and one-step adversarial attacks (\textit{e.g.}, FGSM \cite{goodfellow2014explaining}) \cite{xiao2022stability,jin2022enhancing,guzman2022cross,li2022subspace,zhang2022revisiting,jia2022boosting}. 

\begin{figure}[ht]
    \centering
    \begin{subfigure}
        \centering
        \includegraphics[width=0.85\linewidth]{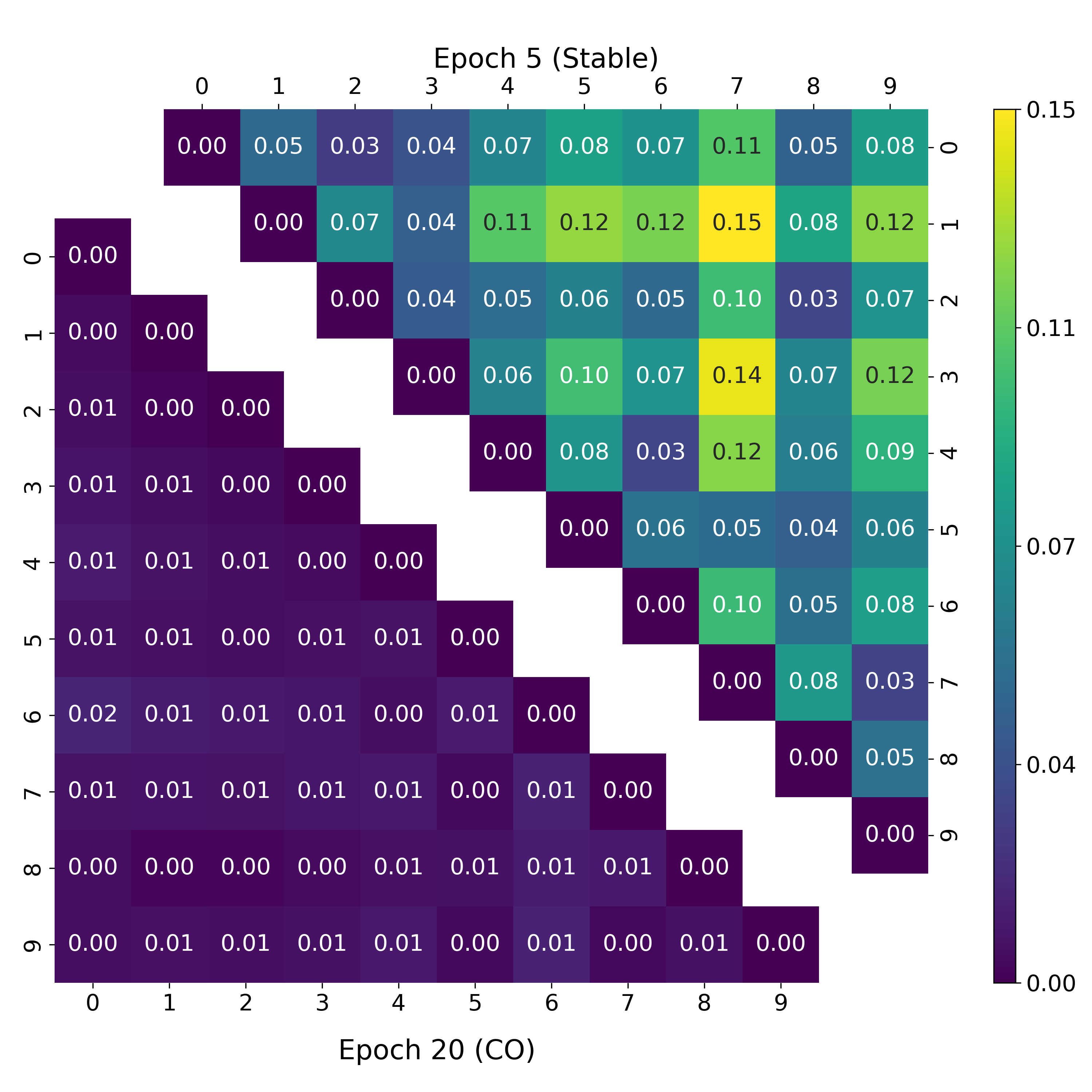}
        \vspace{-4mm}
        \caption*{(a) Distance confusion matrix}
        \label{fig:cm}
    \end{subfigure}
    
    \vspace{0.5em}
    
    \begin{subfigure}
        \centering
        \includegraphics[width=1\linewidth]{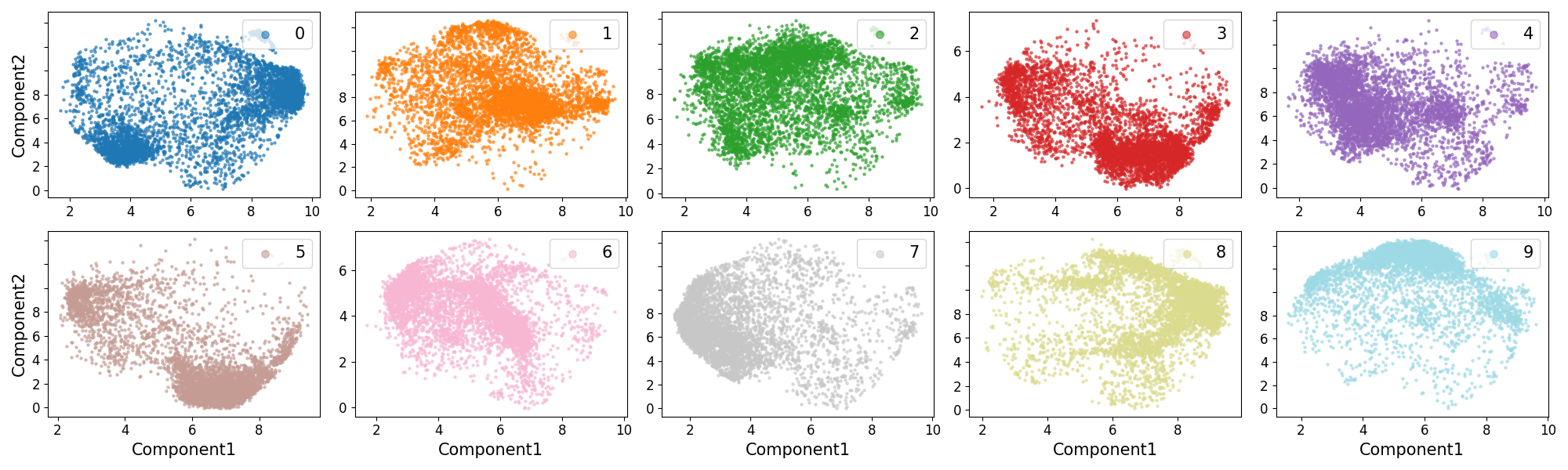}
        \vspace{-3mm}
        \caption*{(b) UMAP distribution of \(\delta_\text{sign}\) in the stably trained model}
        \label{fig:umap-stable}
    \end{subfigure}
    
    \vspace{0.5em}
    
    \begin{subfigure}
        \centering
        \includegraphics[width=1\linewidth]{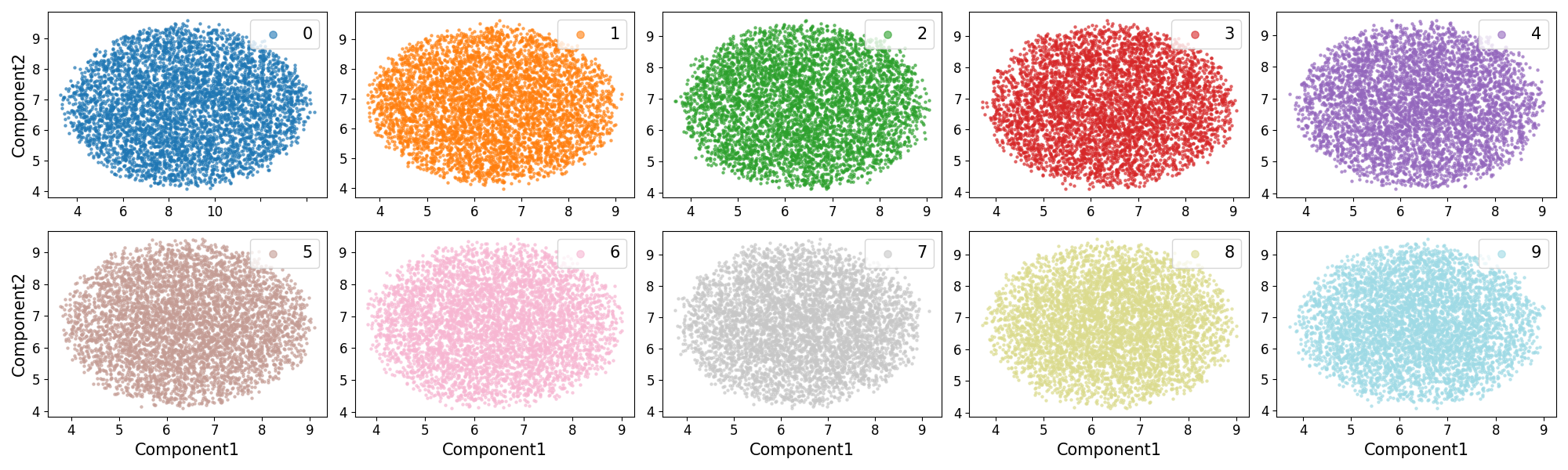}
        \vspace{-3mm}
        \caption*{(c) UMAP distribution of \(\delta_\text{sign}\) in the CO-affected model}
        \label{fig:umap-co}
    \end{subfigure}
    
    \caption{Distance matrix and UMAP visualizations under FGSM-RS. Small distances imply reduced separation between class distributions.}
    \label{fig:combined}
\end{figure}

Compared to PGD-based methods,  FGSM-based methods such as FGSM-RS \cite{Wong2020} and FGSM-MEP \cite{jia2022prior}, also called FAT \cite{tong2024taxonomy,jia2024revisiting}, are computationally efficient \cite{jia2024fast}.
Given the initial perturbation $\delta_0\in\mathcal{N}(0, \mathbf{I})$, training dataset \(\mathcal{D}\), network $f(\cdot;\theta)$, loss $\mathcal{L}$, step size $\epsilon$, and budget $\xi$, {FGSM-RS} generates adversarial perturbations by Eqs. (\ref{eq1}) and (\ref{eq1sup}),
\begin{equation}\label{eq1}
\delta = \text{clip}_\xi\Bigl(\delta_0 + \epsilon \cdot \delta_{\text{sign}}\Bigr),
\end{equation}
\begin{equation}\label{eq1sup}
\delta_{\text{sign}} = \text{sign}\Bigl(\nabla_{\delta_0}\mathcal{L}\bigl(f(x+\delta_0;\theta), y\bigr)\Bigr),
\end{equation}
and implements adversarial training by Eq. (\ref{eq2}),
\begin{equation}\label{eq2}
\min_{\theta} \mathbb{E}_{x \sim \mathcal{D}} \mathcal{L}\Bigl( f\bigl( x + \delta; \theta \bigr), y \Bigr).
\end{equation}
In contrast to FGSM-RS, FGSM-MEP constructs \(\delta_0\) by leveraging the momentum accumulated from adversarial perturbations computed over preceding epochs. It also incorporates a prediction regularization during minimization, expressed as
\begin{equation}\label{eq9}
\mathcal{R}_{\text{pred}} = \beta\|f(x+\delta) - f(x+\delta_0)\|_2^2.
\end{equation}

However, FAT approaches may suffer from CO \cite{de2022make,rice2020overfitting,jia2024improving}.
To address this issue, various strategies have been proposed \cite{zareapoor2024rethinking}, such as gradient alignment \cite{andriushchenko2020understanding}, convergence smoothness \cite{zhao2023fast}, zero-gradient clipping \cite{golgooni2023zerograd}, bi-level optimization \cite{wang2024preventing} and feature activation consistency \cite{zhao2024catastrophic}. Recently, He et al. \cite{he2023investigating} assume that adversarial perturbations embed self-information and argues that models acquire this information through a separate pathway. Then, Zhao et al. \cite{zhao2024catastrophic} introduce a strong regularization term that enforces consistency between predictions on clean and adversarial examples, thereby suppressing the emergence of the adversarial pathway. Lin et al. \cite{LinRevealing} leverage both weight-level and example-level adversarial perturbations to enhance training stability by enforcing weight robustness.

\mydeleted{In contrast, this work explains CO through the lens of backdoor, unifying CO, backdoor attacks, and unlearnable tasks under a common interpretation of trigger overfitting. 
Within this perspective, this work introduces several backdoor-inspired mitigation strategies that enable models to break free from CO and also suppress its occurrence.
Particularly, unlike existing methods that add regularization constraints to the prediction space or feature space for suppressing CO, this work realizes this by adjusting the distribution of model weights.
In addition, while both this work and \cite{LinRevealing} identify weight anomalies, \cite{LinRevealing} aims to construct robust weights, whereas our focus is solely on suppressing weight outliers. We observe that enforcing weight robustness will limit the model’s ability to fit clean samples.
Moreover, although our weight outlier suppression constraint and the strong regularization in \cite{zhao2024catastrophic} are both effective in preventing the adversarial pathway, the latter is often affected by the misclassification on clean examples.}
\myadded{In contrast, this work explains CO through the lens of backdoor, unifying CO, backdoor attacks, and unlearnable tasks under a common interpretation of trigger overfitting. 
Within this perspective, we introduce several backdoor-inspired mitigation strategies that not only enable models to break free from CO but also suppress its occurrence.
Particularly, unlike existing methods that suppress CO by adding regularization constraints to the prediction or feature space, this work adjusts the distribution of model weights.
Furthermore, while both this work and that of Lin et al. \cite{LinRevealing} identify weight anomalies, their objective is to construct robust weights, whereas ours is specifically to suppress weight outliers. We observe that enforcing overall weight robustness will limit the model’s ability to fit clean samples.
Moreover, although both our weight outlier suppression constraint and the strong regularization proposed by \cite{zhao2024catastrophic} are effective at preventing adversarial pathways, the latter is often compromised by misclassifications on clean examples. Additionally, it does not explain why adversarial paths emerge and override data paths.
This work addresses this question from the perspective of trigger overfitting.}

\subsection{Backdoor-related tasks}
This section covers both backdoor attacks and unlearnable tasks, the latter representing a transferable application of backdoor attacks.

Backdoor attacks pose a serious threat to the security of deep neural networks. Early works such as BadNets \cite{gu2019badnets} and TrojanNN \cite{liu2018trojaning} demonstrate that inserting poisoned samples with static triggers during training can cause targeted misclassification. However, such attacks are often detectable due to their reliance on fixed and conspicuous patterns.

To improve stealth and generalizability, a range of trigger designs have been proposed \cite{li2021invisible,zeng2021rethinking,chen2025fsba}. For instance, spatial transformations \cite{nguyen2021wanet}, image blending \cite{chen2017targeted}, and frequency-domain perturbations \cite{wang2022invisible,zeng2021rethinking} aim to create imperceptible or input-adaptive triggers. Others have explored learnable or sample-specific backdoor strategies \cite{li2021invisible, doan2021lira} to improve trigger effectiveness and bypass detection.
More recently, attention has turned to contrastive and self-supervised learning frameworks. CTRL \cite{li2023embarrassingly} shows that even models trained without labels are susceptible to backdoor insertion. 

Beyond implanting malicious behaviors into trained models, backdoor techniques have been adapted to unlearnable tasks, which aim to impair a model’s generalization on clean data. Huang et al. \cite{huang2021unlearnable} introduce sample-wise and class-wise perturbations—similar in nature to triggers—into all training samples, inducing overfitting and preventing the learning of useful representations. However, their effectiveness diminishes under different training settings or datasets. To improve transferability, Ren et al. \cite{ren2022transferable} propose a Classwise Separability Discriminant strategy that enhances linear separability. Zhang et al. \cite{zhang2023unlearnable} generate label-agnostic unlearnable examples via cluster-wise perturbations. Liu et al. \cite{liu2024multimodal} extend protection to multimodal contrastive learning. 

Notably, Qin et al. \cite{qin2023learning} show that adversarial augmentations can mitigate unlearnable-example attacks, motivating Fu et al. \cite{fu2022robust} to design robust unlearnable examples against adversarial learning. Furthermore, Ye et al. \cite{ye2024ungeneralizable} present ungeneralizable samples that can only be learned by specific networks, and Li et al. \cite{li2025detecting} develop methods to detect and corrupt convolution-based unlearnable examples.

This work considers both standard backdoor attacks and unlearnable tasks, and investigates their connections to CO.



\begin{table*}[t]
  \caption{\myadded{Description of symbols.}}
  \label{table0}
  \setlength{\tabcolsep}{0.11cm}
  \centering
  \begin{tabular}{ll}
    \toprule
    \textbf{Symbols}& \textbf{Description}   \\
    \midrule
    $\delta; \delta_0; \delta_\text{sign}; \delta_\text{mom}$&Adversarial perturbation; Initial perturbation; Perturbation direction; Momentum-based perturbation\\
    $\theta; \theta_\text{adv}; \theta_\text{data}; \theta_\text{CO}$& Model parameters; Adversarial path parameters; Data path parameters; CO-affected model parameters\\
    $\xi; \xi_\text{T}; \xi_\text{E}$& Maximum perturbation budget; Maximum perturbation budget during training; Maximum perturbation budget during evaluation\\
    $\eta; \alpha; \epsilon$& Hyperparameters; Step size\\
    \bottomrule
  \end{tabular}
\end{table*}

\section{ CO and backdoor-related tasks}\label{comparison}
In this section, we first present the motivation for interpreting CO through the lens of backdoor. We then provide a comparative analysis highlighting the similarities and differences between CO and backdoor-related tasks.
Table \ref{table0} summarizes the notation used in this paper for quick reference.
\subsection{Motivation of explaining CO from backdoor}
\label{gen_inst}

Prior FAT work \cite{zhao2024catastrophic} has proposed two key hypotheses regarding CO: (i) CO-affected models can be decomposed into distinct adversarial and data branches; (ii) the adversarial branch exhibits feature overriding for adversarial inputs. In this paper, we systematically explore these hypotheses by analyzing the behaviors of FGSM-RS and FGSM-MEP.

\begin{figure}[t]
    \centering
    \begin{subfigure}
        \centering
        \includegraphics[width=0.85\linewidth]{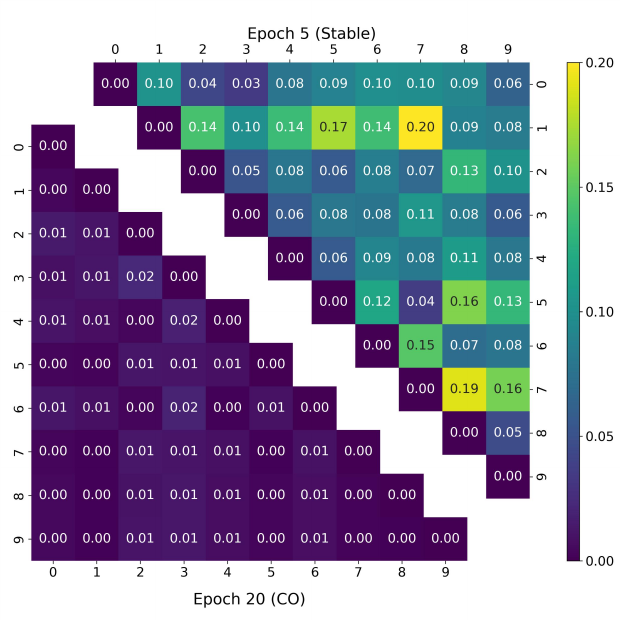}
        \vspace{-4mm}
        \caption*{(a) Distance confusion matrix}
        \label{fig:cm2}
    \end{subfigure}
    
    \vspace{0.5em}
    
    \begin{subfigure}
        \centering
        \includegraphics[width=1\linewidth]{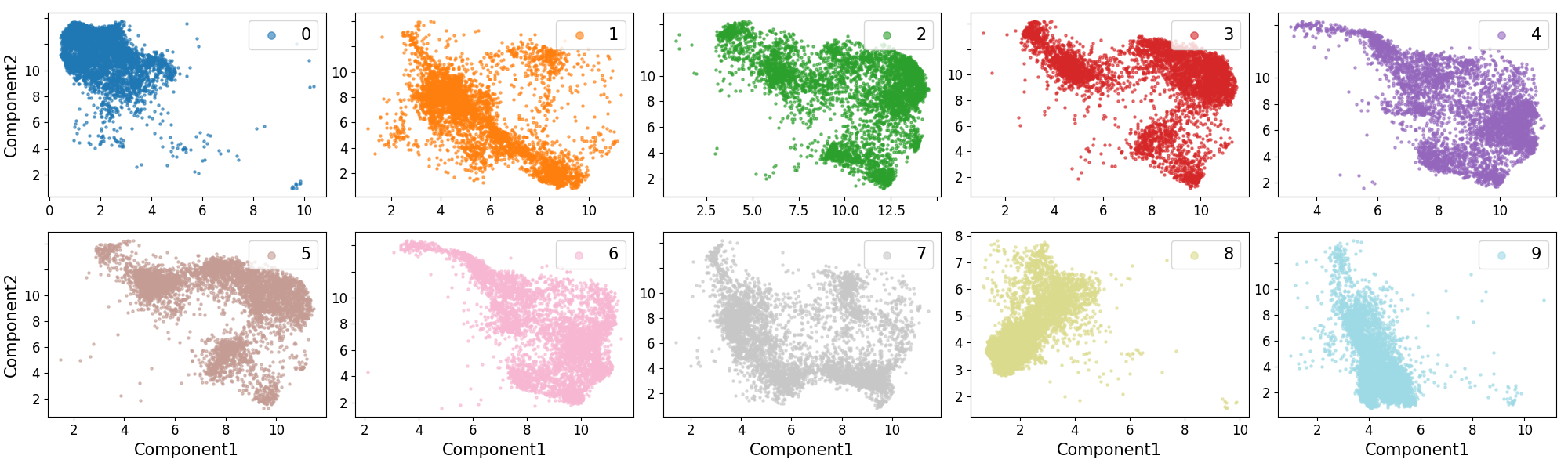}
        \vspace{-3mm}
        \caption*{(b) UMAP distribution of \(\delta_\text{sign}\) in the stably trained model}
        \label{fig:umap-stable2}
    \end{subfigure}
    
    \vspace{0.5em}
    
    \begin{subfigure}
        \centering
        \includegraphics[width=1\linewidth]{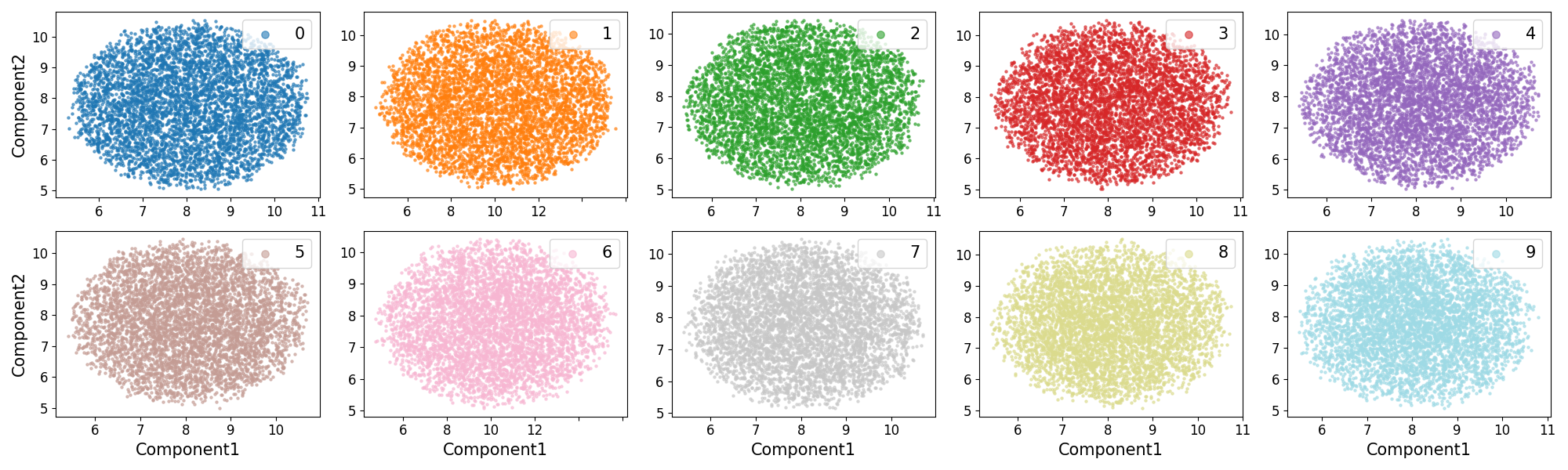}
        \vspace{-3mm}
        \caption*{(c) UMAP distribution of \(\delta_\text{sign}\) in the CO-affected model}
        \label{fig:umap-co2}
    \end{subfigure}
    
    \caption{Distance matrix and UMAP visualizations under FGSM-MEP. Small distances imply reduced separation between class distributions.}
    \label{fig:combined2}
\end{figure}

\textit{Pathway division}: In the absence of the adversarial pathway, the primary distribution of adversarial perturbations \(\delta_\text{sign}\) should be class-separable.
To verify this, we compute the inter-class Wasserstein distance based on $\delta_{\mathrm{sign}}$.
The experiments are conducted on CIFAR-10 \cite{krizhevsky2009learning} using a ResNet18 backbone \cite{he2016deep}, with the maximum perturbation budget of 16/255 and the learning rate of 0.1.
The results in Figure~\ref{fig:combined}(a) and Figure~\ref{fig:combined2}(a) reveal that the stable model (epoch 5) maintains distinct inter-class distances, while the CO-affected model (epoch 20) exhibits distribution overlap (distance $0$). These observations are further supported by Uniform Manifold Approximation and Projection (UMAP) distribution \cite{mcinnes2018umap} in Figures~\ref{fig:combined} and \ref{fig:combined2}, showing class-separable patterns in stable models versus complete overlap under CO-affected models.\footnote{UMAP is a dimensionality reduction technique used for visualizing primary and high-dimensional features.}
{Namely, the CO-affected model should contain an additional adversarial branch.}
We formally decompose the network parameters as $\theta = \{\theta_{\text{adv}}, \theta_{\text{data}}\}$, corresponding to the adversarial and data pathways, respectively. The lack of feature discrimination in $\delta_{\text{sign}}$ suggests that the adversarial pathway dominates gradient backpropagation, expressed as
\begin{equation}\label{eq4}
\nabla_{\delta_0}\mathcal{L}(f(x+\delta_0;\theta), y) \approx \nabla_{\delta_0}\mathcal{L}(f(x+\delta_0;\theta_{\text{adv}}), y).
\end{equation}

\begin{table*}[t]
  \caption{Comparison between backdoor-related tasks and CO.}
  \label{table1}
  \setlength{\tabcolsep}{0.11cm}
  \centering
  \begin{tabular}{llll}
    \toprule
    & \multicolumn{2}{c}{\textbf{Backdoor Attack}}   & \multirow{2}{*}{\textbf{Catastrophic Overfitting (CO)}}    \\
    & Default & Unlearnable & \\
    \midrule
     \multirow{2}{*}{1} & High accuracy for triggered inputs \(x_\text{trigger}\)& High accuracy for perturbed inputs \(x_\text{perturbed}\)& {High accuracy for adversarial examples \(x + \text{FGSM}(x)\)}\\
    & (classified as the predefined class) & (classified as the truth class)& (classified as the truth class)\\
    \midrule[0.05pt]
    \multirow{2}{*}{2} & \multirow{2}{*}{High accuracy for \(x\)} &{Low accuracy for \(x\) }& FGSM-RS: Moderate accuracy for \(x\)
    \\
    &&From low to high as the perturbation size increases& FGSM-MEP: Higher accuracy than FGSM-RS for \(x\) \\
    \midrule[0.05pt]
    3 & Low accuracy for attacks        & Low accuracy for attacks & Low accuracy for attacks except FGSM\\
    \bottomrule
  \end{tabular}
\end{table*}

\begin{figure}[t]
    \centering
    \begin{subfigure}
        \centering
        \includegraphics[width=\linewidth]{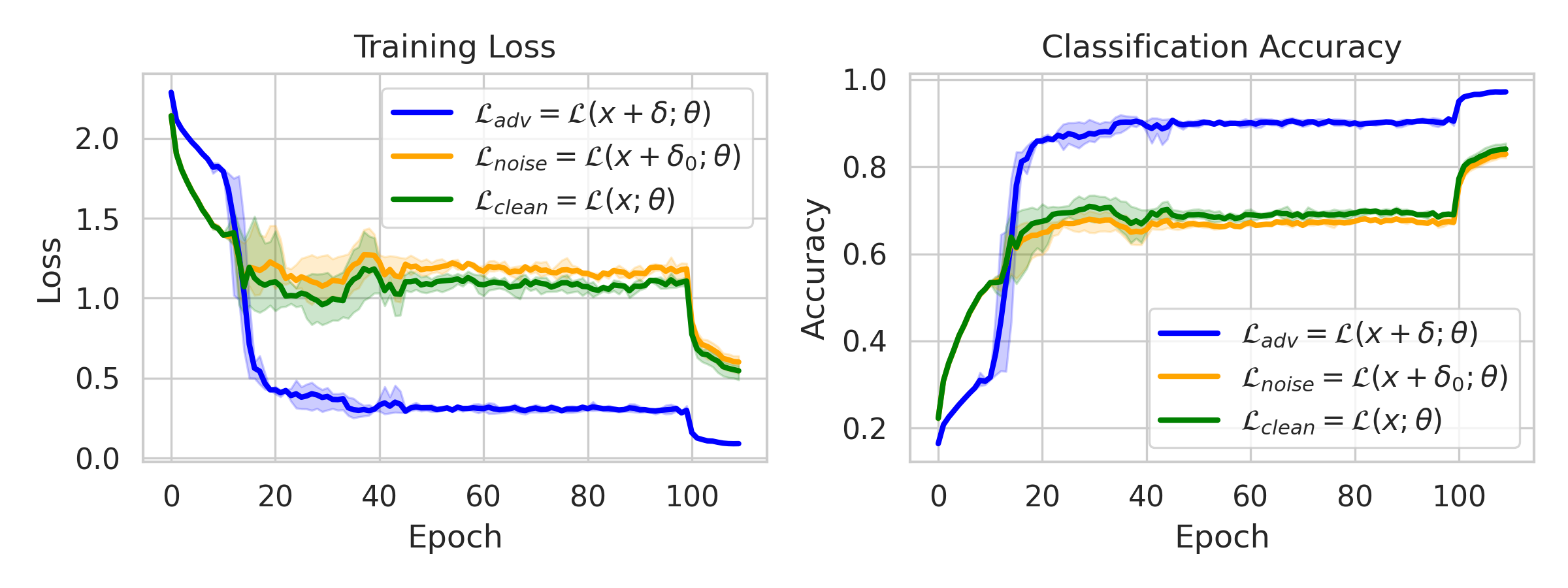}
        \vspace{-5mm}
        \caption*{(a) FGSM-RS (CO in 13th Epoch)}
        \label{fig:loss-rs}
    \end{subfigure}

    \begin{subfigure}
        \centering
        \includegraphics[width=\linewidth]{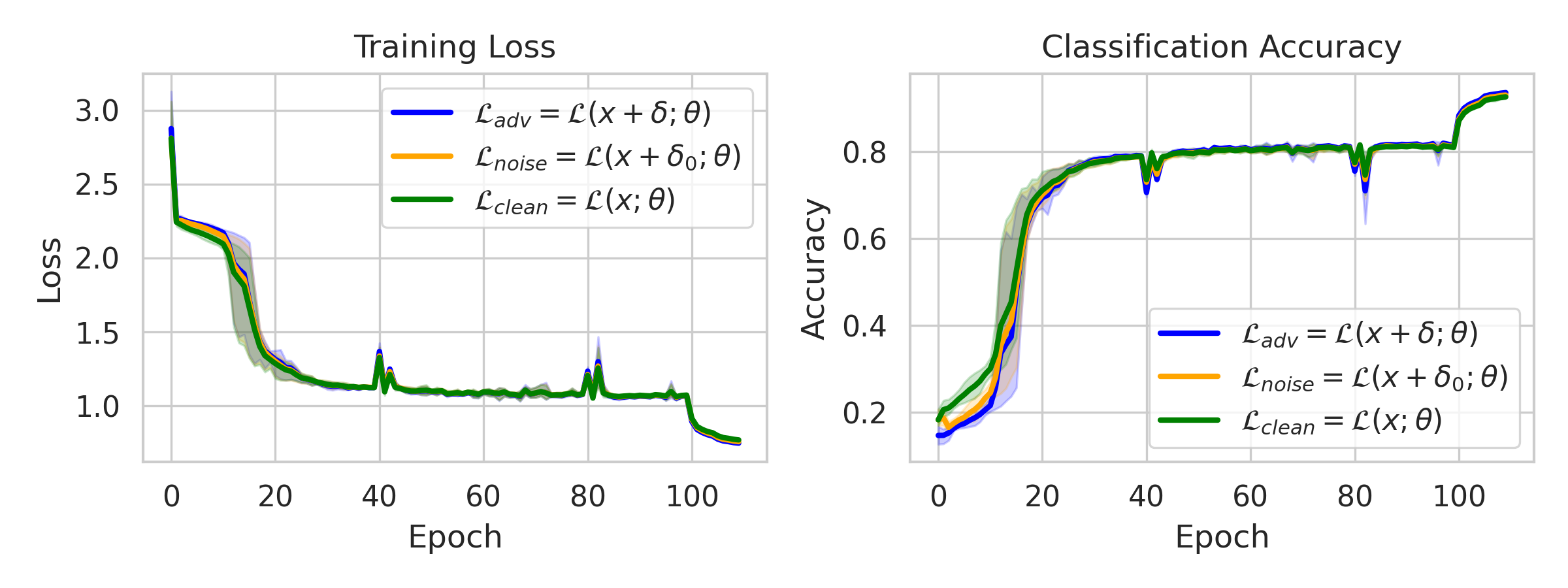}
        \vspace{-5mm}
        \caption*{(b) FGSM-MEP (CO in 12th Epoch)}
        \label{fig:loss-mep}
    \end{subfigure}
    \caption{Training dynamics of existing FAT methods.}
    \label{fig:loss-combined}
\end{figure}

\textit{Diverse forward predictions instead of feature overriding.}
To analyze the forward prediction behavior of FAT models, we visualize the training dynamics of diverse methods on CIFAR-10 with ResNet-18 in Figure~\ref{fig:loss-combined}.
After CO, we observe:
\begin{equation}\label{eq5}
\begin{aligned}
    &\mathbb{E}_{x\sim\mathcal{D}_{\text{train}}} \text{ACC}(x+\delta) \\
    &\geq \mathbb{E}_{x\sim\mathcal{D}_{\text{train}}} \text{ACC}(x+\delta_0) \approx \mathbb{E}_{x\sim\mathcal{D}_{\text{train}}} \text{ACC}(x) \gg 0.
\end{aligned}
\end{equation}
1) Eq.~(\ref{eq5}) implies that CO-affected models exhibit higher or comparable classification accuracy on adversarial examples than on clean \(x\) and initially perturbed samples \(x+\delta_0\).
2) Meanwhile, CO-affected models remain classifiable to \(x\) and \(x+\delta_0\).
These observations indicate that, during forward inference, the CO-affected model may integrate features from both data and adversarial pathways,
\begin{equation}\label{eq6}
\begin{aligned}
&\min_\theta \mathcal{L}(f(x+\delta;\theta),y) \approx\\ &\min_{\theta_{\text{adv}}, \theta_{\text{data}}} \Bigl[\mathcal{L}(f(x+\delta;\theta_{\text{adv}}),y) + \mathcal{L}(f(x+\delta;\theta_{\text{data}}),y)\Bigr],
\end{aligned}
\end{equation}
or directly utilize the data pathway,
\begin{equation}\label{eq62}
\min_\theta \mathcal{L}(f(x+\delta;\theta),y) \approx\min_{\theta_{\text{data}}} \mathcal{L}(f(x+\delta;\theta_{\text{data}}),y).    
\end{equation}


\newcommand{\shadedcell}[2]{\multicolumn{1}{>{\cellcolor{#1}}c}{#2}}
\begin{table*}[t]
\caption{Comparison between various FAT methods. 
Methods marked with \(\dagger\) and  \(\ddagger\) utilize Eq. (\ref{eq7}), with the parameter \(\alpha\) set to 1e$^{-2}$ and 1, respectively.
`Best' and `Final' refer to the evaluation results of the model with the best PGD-10 performance and the final model, respectively. 
`Clean' denotes the classification accuracy for clean examples.
`Perturbed' means that we employ initially perturbed samples. FGSM, PGD, C\&W~\cite{carlini2017towards} and AA~\cite{Francesco2020} are adversarial attacks for evaluating robustness. 
`CO' signifies whether the trained model falls into catastrophic overfitting during FAT.}
  \label{table2}
  \centering
  \setlength{\tabcolsep}{0.15cm} 
  \begin{tabular}{c l c c c c c c c c c c}
    \toprule
    \multicolumn{1}{c}{\textbf{Net\&Dataset}} &
    \multicolumn{1}{c}{\textbf{Methods}} & 
    \multicolumn{1}{c}{ } &
    \textbf{Clean}$\uparrow$ & \textbf{Perturbed}$\uparrow$ & \textbf{FGSM}$\uparrow$ & \textbf{PGD10}$\uparrow$ & \textbf{PGD20}$\uparrow$ & \textbf{PGD50}$\uparrow$ & \textbf{C\&W}$\uparrow$ & \textbf{AA}$\uparrow$ & 
    \multicolumn{1}{c}{\textbf{CO}} \\
    \midrule
    &\multirow{2}{*}{{\cellcolor{white}FGSM-RS \cite{Wong2020}}} 
      & \shadedcell{gray!20}{best}  & \shadedcell{gray!20}{54.69} & \shadedcell{gray!20}{54.64} & \shadedcell{gray!20}{34.31} & \shadedcell{gray!20}{26.01} & \shadedcell{gray!20}{19.77} & \shadedcell{gray!20}{17.85} & \shadedcell{gray!20}{18.41} & \shadedcell{gray!20}{12.63} & \multicolumn{1}{c}{\cellcolor{white}\multirow{2}{*}{\ding{52}}} \\
      && final & \shadedcell{white}{80.80} & \shadedcell{white}{83.16} & \shadedcell{white}{76.62} & \shadedcell{white}{0.00}  & \shadedcell{white}{0.00}  & \shadedcell{white}{0.00}  & \shadedcell{white}{0.00}  & \shadedcell{white}{0.00}  &  \\
    \multirow{2}{*}{{\cellcolor{white} ResNet18 \cite{he2016deep}}}&\multirow{2}{*}{{\cellcolor{white}FGSM-RS\(^{\dagger}\)\cite{Wong2020}}} 
     & \shadedcell{gray!20}{best}  & \shadedcell{gray!20}{73.55} & \shadedcell{gray!20}{73.05} & \shadedcell{gray!20}{44.00} & \shadedcell{gray!20}{34.20} & \shadedcell{gray!20}{23.99} & \shadedcell{gray!20}{20.78} & \shadedcell{gray!20}{22.74} & \shadedcell{gray!20}{14.63} & \multicolumn{1}{c}{\cellcolor{white}\multirow{2}{*}{\ding{55}}} \\
      && final & \shadedcell{white}{73.34} & \shadedcell{white}{73.05} & \shadedcell{white}{44.15} & \shadedcell{white}{33.99} & \shadedcell{white}{24.04} & \shadedcell{white}{21.00} & \shadedcell{white}{22.61} & \shadedcell{white}{14.76} &  \\
    \multirow{2}{*}{{\cellcolor{white} CIFAR10 \cite{krizhevsky2009learning}}} &\multirow{2}{*}{{\cellcolor{white}FGSM-MEP\cite{jia2022prior}}} 
      & \shadedcell{gray!20}{best}  & 
      \shadedcell{gray!20}{30.83}& \shadedcell{gray!20}{30.58}& \shadedcell{gray!20}{25.69}& \shadedcell{gray!20}{23.09}& \shadedcell{gray!20}{21.67}& \shadedcell{gray!20}{21.49}& \shadedcell{gray!20}{20.35}& \shadedcell{gray!20}{18.78}& \multicolumn{1}{c}{\cellcolor{white}\multirow{2}{*}{\ding{52}}} \\
& &final&\textbf{88.07}& \textbf{88.34} & \textbf{80.03} & 11.40& 6.75& 3.74& 3.48& 0.08\\

    &\multirow{2}{*}{{\cellcolor{white}FGSM-MEP\(^{\ddagger}\)\cite{jia2022prior}}} 
      & \shadedcell{gray!20}{best}  
& \shadedcell{gray!20}{59.73}& \shadedcell{gray!20}{59.05}& \shadedcell{gray!20}{42.27}& \shadedcell{gray!20}{\textbf{37.39}}& \shadedcell{gray!20}{\textbf{32.43}} & \shadedcell{gray!20}{\textbf{31.60}}& \shadedcell{gray!20}{\textbf{25.97}}& \shadedcell{gray!20}{\textbf{22.22}}& \multicolumn{1}{c}{\cellcolor{white}\multirow{2}{*}{\ding{55}}} \\
&& final &59.69& 59.09& 42.02& 37.25 & 32.20 & 30.99& 25.67& 22.19\\

\midrule

    &\multirow{2}{*}{{\cellcolor{white}FGSM-RS \cite{Wong2020}}} 
      & \shadedcell{gray!20}{best} &\shadedcell{gray!20}{56.38} & \shadedcell{gray!20}{56.22} & \shadedcell{gray!20}{33.95} & \shadedcell{gray!20}{28.21} & \shadedcell{gray!20}{21.71} & \shadedcell{gray!20}{19.66} & \shadedcell{gray!20}{18.81} & \shadedcell{gray!20}{13.49}&\multirow{2}{*}{\ding{52}}\\
      && final &81.22 & 80.56 & 80.33 & 0.00 & 0.00 & 0.00 & 0.00 & 0.00\\
    \multirow{2}{*}{{\cellcolor{white}PreActResNest18 \cite{he2016identity}}}&\multirow{2}{*}{{\cellcolor{white}FGSM-RS\(^{\dagger}\) \cite{Wong2020}}} 
      & \shadedcell{gray!20}{best} & \shadedcell{gray!20}{72.71} &  \shadedcell{gray!20}{72.21} & \shadedcell{gray!20}{45.45} &  \shadedcell{gray!20}{34.94} &  \shadedcell{gray!20}{25.11} &  \shadedcell{gray!20}{22.38} & \shadedcell{gray!20}{22.21} &  \shadedcell{gray!20}{14.78}&\multirow{2}{*}{\ding{55}}  \\
      && final &72.68 & 72.49 & 44.78 & 34.76 & 25.01 & 21.90 & 22.22 & 14.82 \\
    \multirow{2}{*}{{\cellcolor{white}CIFAR10 \cite{krizhevsky2009learning}}}&\multirow{2}{*}{{\cellcolor{white}FGSM-MEP \cite{jia2022prior}}} 
    & \shadedcell{gray!20}{best}& \shadedcell{gray!20}{41.81} &  \shadedcell{gray!20}{41.78} &  \shadedcell{gray!20}{29.19} &  \shadedcell{gray!20}{26.87} & \shadedcell{gray!20}{ 24.26 }&  \shadedcell{gray!20}{23.89} &  \shadedcell{gray!20}{18.42} &  \shadedcell{gray!20}{17.08} &\multirow{2}{*}{\ding{52}}\\
      && final&\textbf{88.66} & \textbf{89.43} & \textbf{82.76} & 11.17 & 6.82 & 4.09 & 2.84 & 0.01\\
    &\multirow{2}{*}{{\cellcolor{white}FGSM-MEP\(^{\ddagger}\) \cite{jia2022prior}}} 
      & \shadedcell{gray!20}{best} &  \shadedcell{gray!20}{59.29} &  \shadedcell{gray!20}{58.28} &  \shadedcell{gray!20}{42.07} &  \shadedcell{gray!20}{\textbf{36.69}} &  \shadedcell{gray!20}{\textbf{31.94}} &  \shadedcell{gray!20}{\textbf{30.74}} &  \shadedcell{gray!20}{25.50} &  \shadedcell{gray!20}{\textbf{21.40}} &\multirow{2}{*}{\ding{55}}\\
&& final &59.61 & 58.90 & 41.85 & 36.29 & 31.14 & 30.07 & \textbf{25.54} & 21.26\\

\midrule

    &\multirow{2}{*}{{\cellcolor{white}FGSM-RS \cite{Wong2020}}} 
      & \shadedcell{gray!20}{best} &\shadedcell{gray!20}{30.40} & \shadedcell{gray!20}{30.28} & \shadedcell{gray!20}{14.78} & \shadedcell{gray!20}{11.74} & \shadedcell{gray!20}{9.18} & \shadedcell{gray!20}{8.82} & \shadedcell{gray!20}{7.97} & \shadedcell{gray!20}{6.12} &\multirow{2}{*}{\ding{52}} \\
      && final& 39.40 & 52.36 & 47.41 & 0.00 & 0.00 & 0.00 & 0.00 & 0.00\\
    \multirow{2}{*}{{\cellcolor{white} ResNet18 \cite{he2016deep}}}&\multirow{2}{*}{{\cellcolor{white}FGSM-RS\(^{\dagger}\) \cite{Wong2020}}} 
      & \shadedcell{gray!20}{best} & \shadedcell{gray!20}{45.12} & \shadedcell{gray!20}{45.33} & \shadedcell{gray!20}{20.80} & \shadedcell{gray!20}{16.21} & \shadedcell{gray!20}{11.78} & \shadedcell{gray!20}{11.02} & \shadedcell{gray!20}{10.61} & \shadedcell{gray!20}{7.95}&\multirow{2}{*}{\ding{55}}\\
      && final &48.70 & 48.71 & 22.58 & 16.57 & 11.70 & 10.82 & 10.86 & 8.16\\
    \multirow{2}{*}{{\cellcolor{white} CIFAR100 \cite{krizhevsky2009learning}}}&\multirow{2}{*}{{\cellcolor{white}FGSM-MEP \cite{jia2022prior}}} 
    & \shadedcell{gray!20}{best}&\shadedcell{gray!20}{22.18} & \shadedcell{gray!20}{22.17} & \shadedcell{gray!20}{14.51} & \shadedcell{gray!20}{12.57} & \shadedcell{gray!20}{10.95} & \shadedcell{gray!20}{10.73} &\shadedcell{gray!20}{ 8.34} & \shadedcell{gray!20}{7.16} &\multirow{2}{*}{\ding{52}}\\
      && final &\textbf{67.40} & \textbf{67.73} & \textbf{54.74} & 1.33 & 0.60 & 0.27 & 0.31 & 1.00\\
    &\multirow{2}{*}{{\cellcolor{white}FGSM-MEP\(^{\ddagger}\) \cite{jia2022prior}}} 
      & \shadedcell{gray!20}{best}  
&\shadedcell{gray!20}{42.51} & \shadedcell{gray!20}{42.25} & \shadedcell{gray!20}{25.15} & \shadedcell{gray!20}{21.15} &\shadedcell{gray!20}{\textbf{17.28}} & \shadedcell{gray!20}{\textbf{16.71}} & \shadedcell{gray!20}{14.01} & \shadedcell{gray!20}{11.20} &\multirow{2}{*}{\ding{55}}\\
&& final &42.69 & 42.70 & 25.20 & 20.76 & 17.32& 16.59 & \textbf{14.12} & \textbf{11.39}\\

\midrule

        &\multirow{2}{*}{{\cellcolor{white}FGSM-RS \cite{Wong2020}}} 
      & \shadedcell{gray!20}{best} & \shadedcell{gray!20}{27.94} & \shadedcell{gray!20}{27.72} & \shadedcell{gray!20}{13.92} & \shadedcell{gray!20}{11.40} & \shadedcell{gray!20}{9.14} & \shadedcell{gray!20}{8.83} & \shadedcell{gray!20}{7.61} & \shadedcell{gray!20}{6.16} &\multirow{2}{*}{\ding{52}}\\
      && final& 50.30 & 51.80 & 50.00 & 0.00  & 0.00 & 0.00 & 0.00 & 0.00\\
    \multirow{2}{*}{{\cellcolor{white}PreActResNest18 \cite{he2016identity}}}&\multirow{2}{*}{{\cellcolor{white}FGSM-RS\(^{\dagger} \) \cite{Wong2020}}} 
      & \shadedcell{gray!20}{best} &\shadedcell{gray!20}{46.79} & \shadedcell{gray!20}{46.88} & \shadedcell{gray!20}{21.83} & \shadedcell{gray!20}{16.33} & \shadedcell{gray!20}{11.80} & \shadedcell{gray!20}{10.93} & \shadedcell{gray!20}{11.21} & \shadedcell{gray!20}{7.83} &\multirow{2}{*}{\ding{55}} \\
      && final &47.16 & 47.04 & 21.94 & 16.14 & 11.62 & 10.53 & 10.76 & 7.69\\
    \multirow{2}{*}{{\cellcolor{white}CIFAR100 \cite{krizhevsky2009learning}}}&\multirow{2}{*}{{\cellcolor{white}FGSM-MEP \cite{jia2022prior}}} 
    & \shadedcell{gray!20}{best}&\shadedcell{gray!20}{21.21} & \shadedcell{gray!20}{21.02} & \shadedcell{gray!20}{13.61} & \shadedcell{gray!20}{11.79} & \shadedcell{gray!20}{10.22} & \shadedcell{gray!20}{10.01} & \shadedcell{gray!20}{7.72} & \shadedcell{gray!20}{6.75} &\multirow{2}{*}{\ding{52}}\\
      && final&\textbf{67.59} & \textbf{69.12} & \textbf{55.06} & 2.31 & 1.20 & 0.64 & 0.71 & 0.00\\
    &\multirow{2}{*}{{\cellcolor{white}FGSM-MEP\(^{\ddagger}\) \cite{jia2022prior}}} 
      & \shadedcell{gray!20}{best} &\shadedcell{gray!20}{43.22} & \shadedcell{gray!20}{42.96} & \shadedcell{gray!20}{25.91} & \shadedcell{gray!20}{20.84} & \shadedcell{gray!20}{16.77} & \shadedcell{gray!20}{\textbf{16.04}} & \shadedcell{gray!20}{\textbf{14.07}} & \shadedcell{gray!20}{11.18} &\multirow{2}{*}{\ding{55}} \\
&& final &43.22 & 42.95 & 25.99 & \textbf{20.89} & \textbf{16.88} & 16.00 & 14.01 & \textbf{11.20}\\
    \bottomrule
  \end{tabular}
\end{table*}
The validation of pathway decomposition and diverse forward prediction raises a natural question: \textit{Is there an intrinsic connection between CO and backdoor attacks?} 
Table~\ref{table1} shows comparisons between CO and backdoor-related tasks.
Specifically, we hypothesize the existence of a backdoor pathway in the backdoored model. When this pathway is activated, typically by a trigger, feature overriding occurs, leading the model to confidently classify the input into a predefined target class. In the absence of triggers, the model backs to standard prediction behavior and maintains high accuracy on clean samples. Meanwhile, unlearnable tasks \cite{yu2024unlearnable,dolatabadi2024devil,fu2022robust} embed perturbations, functionally similar to backdoor triggers, into all samples while retaining their original labels, causing the model to rapidly overfit to these perturbations (\textit{i.e.} high accuracy for perturbed inputs) and lose generalization capability (\textit{i.e.}, low accuracy for clean examples). Notably, both types remain vulnerable to adversarial attacks, as their training data are not augmented with adversarial examples.
A similar pattern is observed in CO-affected models, which overfit to the specific attack used during training (\textit{e.g.}, FGSM), achieving high accuracy on seen adversarial types while failing to generalize to unseen ones. Taken together, these findings suggest a strong behavioral similarity between CO and backdoor-related tasks.

\subsection{Further analyses between backdoor and CO}\label{sec:comparison}
We further investigate whether FGSM-generated perturbations $\text{FGSM}(x)$ during CO exhibit characteristics similar to backdoor triggers. Specifically, we hypothesize that $\delta_\text{sign}$ contains a universal class-discriminative component, referred to as the UCD trigger $\delta_\text{ucd}$. 

To test this hypothesis, we compute the class-wise expectation of $\delta_\text{sign}$ as defined in Eq.~(\ref{eq8}), and incorporate an auxiliary constraint $\mathcal{L}_\text{aux}$ into the adversarial training objective as specified in Eq.~(\ref{eq7}).
\begin{equation}\label{eq8} \delta_{\text{mom}}^t \leftarrow 0.9\,\delta_{\text{mom}}^t + 0.1\,\mathop{\mathbb{E}}_{x \in \mathcal{B}, y=t} \delta_\text{sign}^{t,*}, \end{equation}
\begin{equation}\label{eq7} \mathcal{L}_\text{aux} = - \alpha \| \delta_\text{sign}^t - \delta_{\text{mom}}^t \|_2 , 
\end{equation}
where $\mathcal{B}$ represents the current batch, $t$ denotes the designated label, and \(\alpha\) is set to 1e$^{-2}$ by default.
The momentum-based term \(\delta_{\text{mom}}^t\) aims to dynamically extract UCD triggers. 
\(\mathcal{L}_\text{aux}\) penalizes class-consistent perturbation features, forcing the model to forget UCD triggers.
Gradients are prevented from propagating into \(\delta_{\text{mom}}^t\) by detaching \(\delta_\text{sign}^t\), denoted as \(\delta_\text{sign}^{t,*}\) in Eq. (\ref{eq8}).
As Table~\ref{table2} shows, \(\mathcal{L}_\text{aux}\) mitigates CO in FGSM-RS and FGSM-MEP across different networks and datasets, confirming the existence of UCD triggers.

\begin{figure*}[t]
    \begin{center}
        \includegraphics[width=1\linewidth]{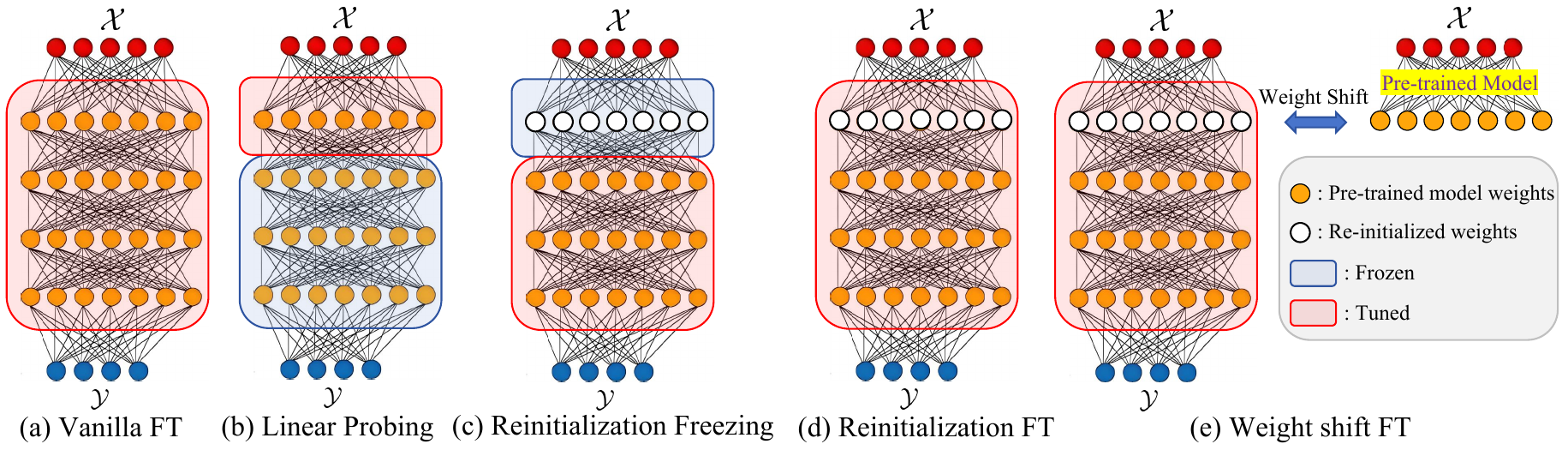}
    \end{center}
  \caption{Backdoor fine-tuning techniques. FT: finetuning.}
    \label{fig9}
\end{figure*}

We next analyze the performance of CO-affected and backdoored models on clean inputs.
The clean classification accuracies across different paradigms are summarized as follows:
\begin{equation}\label{eqsup10}
\begin{aligned}
    \text{ACC}_\text{Ori} &\approx \text{ACC}_\text{SBackdoor} > \text{ACC}_\text{MEP} \gg \text{ACC}_\text{Unlearnable},
\end{aligned}
\end{equation}
where $\text{ACC}_\text{SBackdoor}$, $\text{ACC}_\text{Ori}$, $\text{ACC}_\text{MEP}$, and $\text{ACC}_\text{Unlearnable}$ denote the clean-sample accuracies of models under the standard backdoor, original, CO-affected (trained with FGSM-MEP), and unlearnable settings, respectively.
Notably, $\text{ACC}_\text{MEP}$ falls between $\text{ACC}_\text{SBackdoor}$ and $\text{ACC}_\text{Unlearnable}$, suggesting that CO-affected models preserve a moderate level of clean-data generalization.
In backdoor attacks, the model associates a fixed trigger with a target class while largely preserving its performance on clean data. In contrast, unlearnable tasks embed diverse perturbations to training data without altering labels, causing the model to overfit to these perturbations and lose the ability to accurately classify clean examples.
Notably, in unlearnable tasks, we observe that clean accuracy remains comparable to $\text{ACC}_\text{Ori}$ when the embedded perturbation is weak. As perturbation strength increases, the model’s performance on clean data progressively degrades and collapses once a critical threshold is reached. 
This trend indicates that the higher clean accuracy observed in CO-affected models (\textit{e.g.}, $\text{ACC}_\text{MEP}$) compared to unlearnable tasks ($\text{ACC}_\text{Unlearnable}$) can be attributed to the weaker trigger effect. This is further supported by empirical evidence: as illustrated in Figure~\ref{fig:combined2}, adversarial perturbations in CO-affected models trained with FGSM-MEP are primarily composed of non-discriminative features, with UCD triggers contributing only marginally.

By verifying the existence of UCD triggers in adversarial perturbations generated for CO-affected models, and by analyzing the differences in clean accuracy across techniques, we argue that CO, standard backdoor attacks, and unlearnable tasks can be understood within a unified perspective of trigger overfitting.
\textit{More specifically, CO can be viewed as a weaker variant of unlearnable tasks; both can be regarded as transfer applications of backdoor attacks.}

\begin{table*}[t]
\caption{Comparison of various methods on CIFAR-10 using ResNet-18, with FGSM-RS as the baseline.
``Best" and ``Final" denote results from the checkpoint with the best PGD-10 accuracy and the final epoch, respectively. 
`Clean' denotes the classification accuracy for clean examples.
`Perturbed' means that we employ initially perturbed samples.
FGSM, PGD, C\&W, and AA represent different adversarial attacks used for robustness evaluation. ``ST" indicates the number of stable runs out of three ($\circ$: CO occurred; $\star$: stable training).}
  \label{table3}
  \centering
  \setlength{\tabcolsep}{0.25cm} 
  \begin{tabular}{l l c c c c c c c c l}
    \toprule
    \multicolumn{1}{c}{\textbf{Methods}} & 
    \multicolumn{1}{c}{} &
    \textbf{Clean}$\uparrow$ & \textbf{Perturbed}$\uparrow$ & \textbf{FGSM}$\uparrow$ & \textbf{PGD10}$\uparrow$ & \textbf{PGD20}$\uparrow$ & \textbf{PGD50}$\uparrow$ & \textbf{C\&W}$\uparrow$ & \textbf{AA}$\uparrow$ & 
    \multicolumn{1}{c}{\textbf{ST}} \\
    \midrule
    \multirow{2}{*}{{\cellcolor{white}FGSM-RS \cite{Wong2020}}} 
      & \shadedcell{gray!20}{best}  & \shadedcell{gray!20}{54.69} & \shadedcell{gray!20}{54.64} & \shadedcell{gray!20}{34.31} & \shadedcell{gray!20}{26.01} & \shadedcell{gray!20}{19.77} & \shadedcell{gray!20}{17.85} & \shadedcell{gray!20}{18.41} & \shadedcell{gray!20}{12.63} &\multicolumn{1}{c}{\cellcolor{white}\multirow{2}{*}{\(\circ\)\(\circ\)\(\circ\)}} \\
      & final & \shadedcell{white}{80.80} & \shadedcell{white}{83.16} & \shadedcell{white}{76.62} & \shadedcell{white}{0.00}  & \shadedcell{white}{0.00}  & \shadedcell{white}{0.00}  & \shadedcell{white}{0.00}  & \shadedcell{white}{0.00}  &  \\
    \midrule
      \multirow{2}{*}{\bf \cellcolor{white}VFT-CO-Clean \cite{liu2018fine}} 
      & \shadedcell{gray!20}{best}  & \shadedcell{gray!20}{73.49} 
      &\shadedcell{gray!20}{72.87}&\shadedcell{gray!20}{44.94}&\shadedcell{gray!20}{34.81}&\shadedcell{gray!20}{25.32}&\shadedcell{gray!20}{21.87}&\shadedcell{gray!20}{23.00}&\shadedcell{gray!20}{15.19}&\multicolumn{1}{c}{\cellcolor{white}\multirow{2}{*}{\(\star\)\(\star\)\(\star\)}} \\
      & final & \shadedcell{white}{73.79} &73.34&44.48&34.35&24.64&21.75&22.63&15.11\\
    \midrule
    \multirow{2}{*}{{\cellcolor{white}LP-CO-Clean \cite{tomihari2024understanding}}} 
      & \shadedcell{gray!20}{best}  & 
      \shadedcell{gray!20}{61.29}&\shadedcell{gray!20}{60.83}& \shadedcell{gray!20}{37.78}&\shadedcell{gray!20}{30.98}& \shadedcell{gray!20}{23.82}&\shadedcell{gray!20}{22.14}&\shadedcell{gray!20}{21.02}&\shadedcell{gray!20}{15.35}&\multicolumn{1}{c}{\cellcolor{white}\multirow{2}{*}{\(\circ\)\(\circ\)\(\circ\)}}\\
      & final & 78.43 &81.50 &77.59& 0.00& 0.00& 0.00& 0.00& 0.00
      \\
    \midrule
    \multirow{2}{*}{\bf \cellcolor{white}RF-CO-Clean \cite{min23nips}} 
      & \shadedcell{gray!20}{best} &
\shadedcell{gray!20}{69.92}&\shadedcell{gray!20}{70.74} &\shadedcell{gray!20}{43.28}& \shadedcell{gray!20}{34.70}&\shadedcell{gray!20}{25.62}&\shadedcell{gray!20}{22.81}&\shadedcell{gray!20}{ 22.71}&\shadedcell{gray!20}{15.82}&\multicolumn{1}{c}{\cellcolor{white}\multirow{2}{*}{\(\star\)\(\star\)\(\star\)}}\\
      & final &72.82& 72.75& 44.00& 34.04& 25.04& 21.87& 22.45& 15.22 \\
    \midrule
    \multirow{2}{*}{\bf \cellcolor{white}RFT-CO-Clean \cite{min23nips}} 
      & \shadedcell{gray!20}{best}  &\shadedcell{gray!20}{70.92}&\shadedcell{gray!20}{ 69.63}& \shadedcell{gray!20}{43.29}&\shadedcell{gray!20}{ 34.70} &\shadedcell{gray!20}{25.43}&\shadedcell{gray!20}{ 22.83}&\shadedcell{gray!20}{22.66}&\shadedcell{gray!20}{ 15.81}&\multicolumn{1}{c}{\cellcolor{white}\multirow{2}{*}{\(\star\)\(\star\)\(\star\)}}\\
      & final & 72.82 & 72.67 & 44.08 & 34.02 & 25.12 & 21.98 & 22.53 & 15.28 \\
      \midrule
    \multirow{2}{*}{\bf \cellcolor{white}RSFT-CO-Clean \cite{min23nips}} 
      & \shadedcell{gray!20}{best} &\shadedcell{gray!20}{71.79}&\shadedcell{gray!20}{71.51}&\shadedcell{gray!20}{ 43.78}&\shadedcell{gray!20}{ 34.14}&\shadedcell{gray!20}{ 24.82}&\shadedcell{gray!20}{ 21.59}&\shadedcell{gray!20}{ 22.34}&\shadedcell{gray!20}{ 14.89} &\multicolumn{1}{c}{\cellcolor{white}\multirow{2}{*}{\(\star\)\(\star\)\(\star\)}}\\
      & final & 72.17& 71.80&43.50& 33.84&24.38& 21.55& 22.08& 14.93\\
      
    \bottomrule
  \end{tabular}
\end{table*}

\section{Backdoor-inspired mitigation of CO}  
\label{sec:mitigation}
Motivated by previous comparisons, we attempt to employ backdoor defenses to mitigate CO.
We begin by adapting existing fine-tuning methods designed for backdoor removal. Furthermore, inspired by weight poisoning techniques in backdoor, we introduce a supplementary constraint to suppress weight outliers and enhance stability.

\begin{table*}[t]
\caption{Comparison between various techniques, with FGSM-MEP as baseline. ``Best" and ``Final" denote results from the checkpoint with the best PGD-10 accuracy and the final epoch, respectively. 
`Clean' denotes the classification accuracy for clean examples.
`Perturbed' means that we employ initially perturbed samples.
FGSM, PGD, C\&W, and AA represent adversarial attacks for evaluation. ``ST" indicates the number of stable runs out of three ($\circ$: CO occurred; $\star$: stable training).}
  \label{table4}
  \centering
  \setlength{\tabcolsep}{0.25cm} 
  \begin{tabular}{l l c c c c c c c c l}
    \toprule
    \multicolumn{1}{c}{\textbf{Methods}} & 
    \multicolumn{1}{c}{} &
    \textbf{Clean}$\uparrow$ & \textbf{Perturbed}$\uparrow$ & \textbf{FGSM}$\uparrow$ & \textbf{PGD10}$\uparrow$ & \textbf{PGD20}$\uparrow$ & \textbf{PGD50}$\uparrow$ & \textbf{C\&W}$\uparrow$ & \textbf{AA}$\uparrow$ & 
    \multicolumn{1}{c}{\textbf{ST}} \\
    \midrule
    \multirow{2}{*}{{\cellcolor{white}FGSM-MEP \cite{jia2022prior}}} 
      & \shadedcell{gray!20}{best}  & \shadedcell{gray!20}{59.01} & \shadedcell{gray!20}{58.94} & \shadedcell{gray!20}{35.69} & \shadedcell{gray!20}{28.66} & \shadedcell{gray!20}{21.81} & \shadedcell{gray!20}{20.10} & \shadedcell{gray!20}{20.49} & \shadedcell{gray!20}{12.39} &\multicolumn{1}{c}{\cellcolor{white}\multirow{2}{*}{\(\circ\)\(\circ\)\(\circ\)}} \\
      & final & \shadedcell{white}{82.37} & \shadedcell{white}{82.22} & \shadedcell{white}{71.19} & \shadedcell{white}{13.13} & \shadedcell{white}{8.09}  & \shadedcell{white}{3.12}  & \shadedcell{white}{0.89}  & \shadedcell{white}{0.00}  &  \\
    \midrule
    \multirow{2}{*}{\bf \cellcolor{white}VFT-CO-Clean \cite{liu2018fine}} 
      & \shadedcell{gray!20}{best}  &
      \shadedcell{gray!20}{58.43}& \shadedcell{gray!20}{57.93}& \shadedcell{gray!20}{38.18}& \shadedcell{gray!20}{33.71}& \shadedcell{gray!20}{29.32}& \shadedcell{gray!20}{28.44}& \shadedcell{gray!20}{22.93}& \shadedcell{gray!20}{20.16}&\multicolumn{1}{c}{\cellcolor{white}\multirow{2}{*}{\(\star\)\(\star\)\(\star\)}}
      \\
      & final &58.25& 57.75& 38.28& 33.79& 29.39& 28.62& 23.03& 20.15
\\
    \midrule
    \multirow{2}{*}{{\cellcolor{white}LP-CO-Clean \cite{tomihari2024understanding}}} 
      & \shadedcell{gray!20}{best}  &
      \shadedcell{gray!20}{49.03}&\shadedcell{gray!20}{48.42}&\shadedcell{gray!20}{34.41}&\shadedcell{gray!20}{31.18}&\shadedcell{gray!20}{27.67}&\shadedcell{gray!20}{ 27.20}&\shadedcell{gray!20}{ 22.69}&\shadedcell{gray!20}{20.71}&\multicolumn{1}{c}{\cellcolor{white}\multirow{2}{*}{\(\star\)\(\circ\)\(\circ\)}}
      \\
      & final &49.03& 48.55&34.43& 31.16& 27.67& 27.13& 22.74& 20.75 \\
    \midrule
    \multirow{2}{*}{{\cellcolor{white}RF-CO-Clean \cite{min23nips}}} 
      & \shadedcell{gray!20}{best}  &
      \shadedcell{gray!20}{49.03}& \shadedcell{gray!20}{48.50}& \shadedcell{gray!20}{34.57}& \shadedcell{gray!20}{31.15}& \shadedcell{gray!20}{27.63}& \shadedcell{gray!20}{27.10}& \shadedcell{gray!20}{22.73}& \shadedcell{gray!20}{20.73}&\multicolumn{1}{c}{\cellcolor{white}\multirow{2}{*}{\(\star\)\(\circ\)\(\circ\)}}\\
      & final &49.03& 48.45& 34.44& 31.16& 27.55& 27.17& 22.75& 20.78 \\
    \midrule
    \multirow{2}{*}{\bf \cellcolor{white}RFT-CO-Clean \cite{min23nips}} 
      & \shadedcell{gray!20}{best}  & 
      \shadedcell{gray!20}{55.79}& \shadedcell{gray!20}{55.07}& \shadedcell{gray!20}{39.31}& \shadedcell{gray!20}{34.54}& \shadedcell{gray!20}{30.06}& \shadedcell{gray!20}{29.34}& \shadedcell{gray!20}{24.45}& \shadedcell{gray!20}{21.10}&\multicolumn{1}{c}{\cellcolor{white}\multirow{2}{*}{\(\star\)\(\star\)\(\star\)}}\\
      & final &55.74& 55.05& 38.94& 34.29& 30.11& 29.37& 24.31& 21.04 \\
      \midrule
    \multirow{2}{*}{\bf \cellcolor{white}RSFT-CO-Clean \cite{min23nips}} 
      & \shadedcell{gray!20}{best} &\shadedcell{gray!20}{40.13}&\shadedcell{gray!20}{ 39.65}&\shadedcell{gray!20}{29.60}&\shadedcell{gray!20}{25.85}&\shadedcell{gray!20}{24.57}&\shadedcell{gray!20}{24.26}&\shadedcell{gray!20}{21.14}&\shadedcell{gray!20}{19.41} &\multicolumn{1}{c}{\cellcolor{white}\multirow{2}{*}{\(\star\)\(\star\)\(\star\)}}\\
      & final &39.00& 38.64& 28.45& 25.46& 23.22& 22.88& 20.45& 18.50 \\
    \bottomrule
  \end{tabular}
\end{table*}

\begin{figure*}
\centering  
\includegraphics[width=0.8\textwidth]{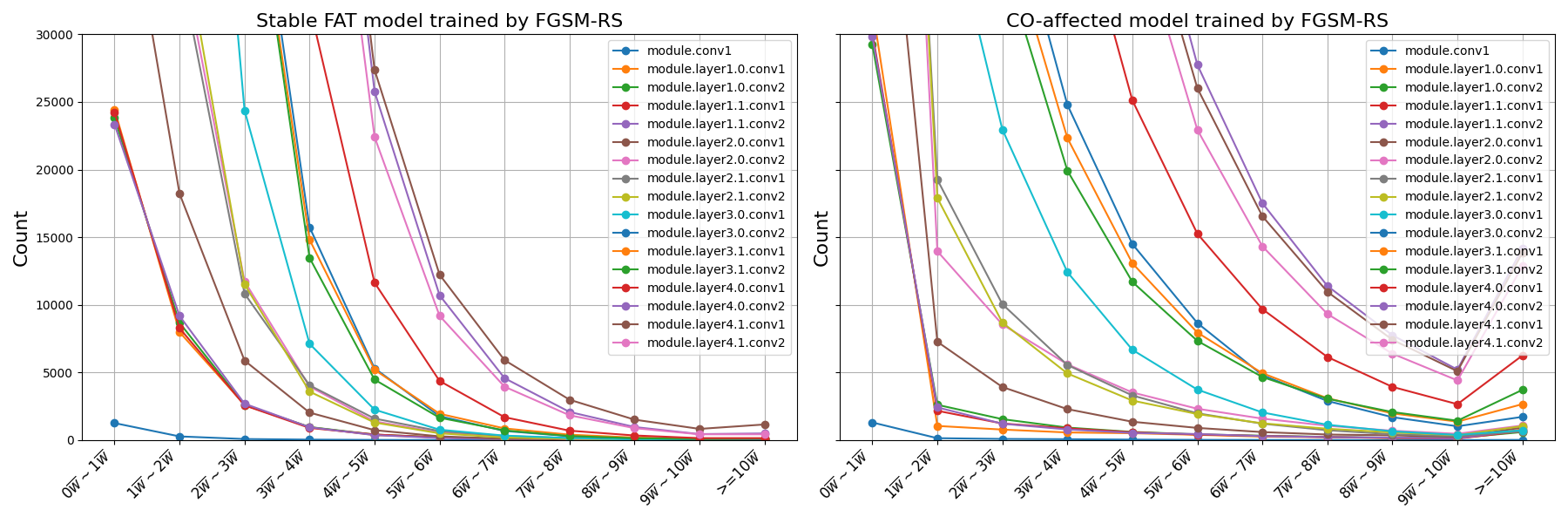} 
\caption{Weight distribution relative to the mean weight. `Count' means the distribution percentage.}  
\label{fig10}  
\end{figure*}

\begin{table*}[t]
\caption{Comparison with existing FAT methods with ResNet18 and CIFAR10. PBD is the state-of-the-art method. 
$\xi_T$ and $\xi_E$ represent the perturbation budget used during the training and evaluation processes. 
``Best" and ``Final" denote results from the checkpoint with the best PGD-10 accuracy and the final epoch, respectively. 
`Clean' denotes the classification accuracy for clean examples.
`Perturbed' means that we employ initially perturbed samples.
``ST" indicates the number of stable runs out of three ($\circ$: CO occurred; $\star$: stable training). For LAP, we directly utilize the results reported in \cite{LinRevealing}.
}
  \label{tableapp7}
  \centering
  \setlength{\tabcolsep}{0.25cm} 
  \begin{tabular}{l l c c c c c c c c l}
    \toprule
    \multicolumn{1}{c}{\(\xi_T = \xi_E = 16/255\)} & 
    \multicolumn{1}{c}{} &
    \textbf{Clean}$\uparrow$ & \textbf{Perturbed}$\uparrow$ & \textbf{FGSM}$\uparrow$ & \textbf{PGD10}$\uparrow$ & \textbf{PGD20}$\uparrow$ & \textbf{PGD50}$\uparrow$ & \textbf{C\&W}$\uparrow$ & \textbf{AA}$\uparrow$ &\textbf{ST}\\
    \midrule
     \multirow{1}{*}{\cellcolor{white} LAP \cite{LinRevealing}} &best & {63.73} & {-} & {-} & {-} &{-} & {-} & {-} & {19.55} &-\\
    \multirow{2}{*}{\cellcolor{white} GradAlign\cite{Andriushchenko2020}} & \shadedcell{gray!20}{best} & \shadedcell{gray!20}{58.17} &\shadedcell{gray!20}{-} & \shadedcell{gray!20}{39.87} & \shadedcell{gray!20}{33.12} & \shadedcell{gray!20}{26.81} & \shadedcell{gray!20}{24.99} & \shadedcell{gray!20}{22.63} & \shadedcell{gray!20}{17.02} &\multicolumn{1}{c}{\cellcolor{white}\multirow{2}{*}{\(\circ\)\(\circ\)\(\circ\)}}\\
    & final & 70.86&- &69.51 & 0.00& 0.00& 0.00& 0.00 & 0.00\\
    
    \multirow{2}{*}{\cellcolor{white} ZeroGrad\cite{Golgooni2021}} &\shadedcell{gray!20}{best}& \shadedcell{gray!20}{74.16}&\shadedcell{gray!20}{-} &\shadedcell{gray!20}{43.96}& \shadedcell{gray!20}{32.67} &\shadedcell{gray!20}{21.98}&\shadedcell{gray!20}{18.37}& \shadedcell{gray!20}{20.76} &\shadedcell{gray!20}{12.07}&\multicolumn{1}{c}{\cellcolor{white}\multirow{2}{*}{\(\star\)\(\star\)\(\star\)}}\\
    & final &75.60 &- & 44.89 &31.77& 20.71 &16.76 &20.09 & 10.87\\
    
    \multirow{2}{*}{\cellcolor{white} NuAT\cite{sriramanan2021towards}} &\shadedcell{gray!20}{best}& \shadedcell{gray!20}{74.62} &\shadedcell{gray!20}{-} & \shadedcell{gray!20}{44.92}& \shadedcell{gray!20}{35.22}& \shadedcell{gray!20}{25.93} &\shadedcell{gray!20}{23.67} &\shadedcell{gray!20}{24.07}& \shadedcell{gray!20}{18.43}&\multicolumn{1}{c}{\cellcolor{white}\multirow{2}{*}{\(\star\)\(\star\)\(\star\)}}\\
    &final & 75.29&- & 45.31& 34.85& 25.58& 23.44& 23.62& 18.06\\
    \midrule
    \multirow{2}{*}{{\cellcolor{white} FGSM-RS \cite{Wong2020}}} 
      & \shadedcell{gray!20}{best}  & \shadedcell{gray!20}{54.69} & \shadedcell{gray!20}{54.64} & \shadedcell{gray!20}{34.31} & \shadedcell{gray!20}{26.01} & \shadedcell{gray!20}{19.77} & \shadedcell{gray!20}{17.85} & \shadedcell{gray!20}{18.41} & \shadedcell{gray!20}{12.63} &\multicolumn{1}{c}{\cellcolor{white}\multirow{2}{*}{\(\circ\)\(\circ\)\(\circ\)}} \\
      & final & \shadedcell{white}{\textbf{80.80}} & \shadedcell{white}{\textbf{83.16}} & \shadedcell{white}{\textbf{76.62}} & \shadedcell{white}{0.00}  & \shadedcell{white}{0.00}  & \shadedcell{white}{0.00}  & \shadedcell{white}{0.00}  & \shadedcell{white}{0.00}  &  \\
    \multirow{2}{*}{{\cellcolor{white} PBD-RS \cite{zhao2024catastrophic}}} 
      & \shadedcell{gray!20}{best}  &\shadedcell{gray!20}{73.32} & \shadedcell{gray!20}{72.96} & \shadedcell{gray!20}{45.03} & \shadedcell{gray!20}{ 35.45} &\shadedcell{gray!20}{ 25.52} & \shadedcell{gray!20}{22.61} & \shadedcell{gray!20}{23.72} & \shadedcell{gray!20}{\textbf{16.21}}&\multicolumn{1}{c}{\cellcolor{white}\multirow{2}{*}{\(\star\)\(\star\)\(\star\)}} \\
      & final & 73.82 & 73.48 & 44.60 & 34.76 & 25.05 & 22.00 & 23.23 & 15.82\\
    \multirow{2}{*}{{\cellcolor{white} Ours($\mathcal{L}_{\text{reg}}$)}} & \shadedcell{gray!20}{best}&
    \shadedcell{gray!20}{74.33}& \shadedcell{gray!20}{74.00}& \shadedcell{gray!20}{46.60}& \shadedcell{gray!20}{\textbf{36.82}}& \shadedcell{gray!20}{\textbf{26.46}}& \shadedcell{gray!20}{\textbf{22.98}}& \shadedcell{gray!20}{\textbf{24.24}}& \shadedcell{gray!20}{15.72}&\multicolumn{1}{c}{\cellcolor{white}\multirow{2}{*}{\(\star\)\(\star\)\(\star\)}}\\
&final&74.84& 74.67&46.81& 36.21& 25.95& 22.70& 24.22&15.70\\
    \midrule
    \multirow{2}{*}{{\cellcolor{white} FGSM-MEP \cite{jia2022prior}}} 
      & \shadedcell{gray!20}{best}  & \shadedcell{gray!20}{59.01} & \shadedcell{gray!20}{58.94} & \shadedcell{gray!20}{35.69} & \shadedcell{gray!20}{28.66} & \shadedcell{gray!20}{21.81} & \shadedcell{gray!20}{20.10} & \shadedcell{gray!20}{20.49} & \shadedcell{gray!20}{12.39} &\multicolumn{1}{c}{\cellcolor{white}\multirow{2}{*}{\(\circ\)\(\circ\)\(\circ\)}} \\
      & final & \shadedcell{white}{\textbf{82.37}} & \shadedcell{white}{\textbf{82.22}} & \shadedcell{white}{\textbf{71.19}} & \shadedcell{white}{13.13} & \shadedcell{white}{8.09}  & \shadedcell{white}{3.12}  & \shadedcell{white}{0.89}  & \shadedcell{white}{0.00}  &  \\
        \multirow{2}{*}{{\cellcolor{white} PBD-MEP \cite{zhao2024catastrophic}}} 
      & \shadedcell{gray!20}{best}  &\shadedcell{gray!20}{64.45} & \shadedcell{gray!20}{63.28} & \shadedcell{gray!20}{45.77} & \shadedcell{gray!20}{39.70} & \shadedcell{gray!20}{34.27} & \shadedcell{gray!20}{32.80} & \shadedcell{gray!20}{27.64} & \shadedcell{gray!20}{22.45} &\multicolumn{1}{c}{\cellcolor{white}\multirow{2}{*}{\(\star\)\(\star\)\(\star\)}} \\
      & final &64.20 & 63.21 & 45.29 & 39.26 & 33.47 & 31.88 & 27.93 & 22.40 \\
    \multirow{2}{*}{{\cellcolor{white} Ours($\mathcal{L}_{\text{reg}}$)}} 
      & \shadedcell{gray!20}{best}  &\shadedcell{gray!20}{65.26}& \shadedcell{gray!20}{64.74}& \shadedcell{gray!20}{46.11}& \shadedcell{gray!20}{\textbf{40.23}}& \shadedcell{gray!20}{\textbf{34.53}}& \shadedcell{gray!20}{\textbf{33.07}}& \shadedcell{gray!20}{27.98}& \shadedcell{gray!20}{\textbf{22.47}}&\multicolumn{1}{c}{\cellcolor{white}\multirow{2}{*}{\(\star\)\(\star\)\(\star\)}} \\
      & final & 65.33& 64.63& 46.19& 40.06& 34.22& 32.73& \textbf{28.13}& 22.21\\
    \bottomrule
  \end{tabular}
\end{table*}

\subsection{Adapted backdoor fine-tuning strategies}  
Once the model $f(x;\theta)$ falls into CO during FAT, we apply one epoch of fine-tuning using the strategies illustrated in Figure~\ref{fig9}, with the details given as follows:
1) VFT \cite{liu2018fine,qin2023revisiting}: Fine-tune all parameters.
2) LP \cite{tomihari2024understanding,ke2024convergence}: Fine-tune first $k$ model layers, 
3) RF \cite{min23nips}: Reinitialize first $k$ layers, freeze them, then fine-tune remaining layers.
4) RFT \cite{min23nips}: Reinitialize first $k$ layers, then fine-tune the whole model.
5) RSFT \cite{min23nips}: Reinitialize first $k$ layers and fine-tune the whole model with an inner product constraint to limit weight shift.

We finetune CO-affected models on clean examples as `*CO-Clean'.
Tables \ref{table3} and \ref{table4} show the experimental results obtained using FGSM-RS and FGSM-MEP, respectively\footnote{Results of finetuning co-affected models on adversarial examples are shown in Appendix.}. Our observations are as follows:
1) Adapted backdoor fine-tuning strategies can also resolve CO in FAT.
2) The clean classification accuracies reported in Table~\ref{table4} are consistently lower than those in Table~\ref{table3}. This degradation may stem from the regularization term $\mathcal{R}_{\text{grad}}$ in Eq.~(\ref{eq9}), which potentially constrains the model’s capacity to fit the clean data distribution. Furthermore, reinitializing a stable FAT process after CO may inherently limit the model's final performance, \textit{e.g.} wrong optimization direction.
3) Notably, even when fine-tuning yields a temporarily stable model, CO tends to reoccur. This is expected, as the goal of backdoor defenses is to transition a trained, backdoored model out of the backdoor state. Overall, fine-tuning strategies designed for backdoor attacks are also applicable to mitigating CO.

\subsection{Weight-poisoning inspired strategy}  
To develop lasting mitigation strategies for CO, we begin by examining training-time backdoor defense techniques~\cite{Kuofeng2023}. These methods typically identify and filter backdoor-infected samples using either one-shot~\cite{Anti2021backdoor} or iterative~\cite{Chen2024AAAI} filtering procedures. In FAT, Zhao et al. \cite{zhao2023fast} demonstrate that applying a convergence-smoothing constraint to a subset of samples with notable convergence divergence can enhance the overall training stability. However, the effectiveness of this method is highly sensitive to biases in sample selection.

Motivated by backdoor approaches with weight poisoning \cite{ShuaiDefending,LiBackdoor,Keita2020Weight}, we examine the weight distributions of both stable and CO-affected models.
Specifically, we compute the mean weight value \(\overline{w}\) for each layer and assess the weight distribution relative to this mean. Figure \ref{fig10} presents experimental results on CIFAR-10 with ResNet-18, showing that CO-affected models exhibit weight outliers—weights that deviate markedly from $\overline{w}$.
Hence, we attempt to address CO by suppressing weight outliers. Unlike backdoor pruning techniques \cite{Yigeneuron,WuDongxian,KunbeiDeepvenom}, which typically clamp weights, we introduce an additional constraint that suppresses weight outliers and prevents the formation of adversarial paths.
\begin{equation}\label{eq12}
\mathcal{L}_{\text{reg}} = \sum_{l\in\mathcal{C}} \sum_{w \in \overline{w}_l} \exp\left(\frac{|w - \overline{w}_l|}{\overline{w}_l + \alpha} - \eta\right) \, |w - \overline{w}_l|,
\end{equation}
\begin{equation}\label{eq122}
\overline{w}_l = \frac{1}{|W_l|}\sum_{w \in W_l} |w|,
\end{equation}
where $\|\cdot\|$ is utilized to obtain the absolute value. \(\mathcal{C}\) denotes the set of convolutional layers, and \(W_l\) represents the set of $l$-th layer weights.
Hyper-parameters \(\eta\) and \(\alpha\) are set to 10 and 10$^{-5}$, respectively.

Table~\ref{tableapp7} reports results on CIFAR-10 with ResNet18, showing that incorporating $\mathcal{L}_{\text{reg}}$ effectively mitigates CO in FAT. 
Additional experiments across diverse datasets, architectures and perturbation budgets are shown in the Appendix, which also bring consistent improvements in mitigating CO.
Overall, the proposed method not only resolves CO but also achieves performance superior to or comparable with the state-of-the-art PBD. 
Unlike PBD, which aligns clean and adversarial predictions at the cost of clean accuracy, and LAT, whose effectiveness is limited by its strict emphasis on robust weights, our method targets only anomalous weights, providing greater flexibility in weight selection.
Additionally, we hypothesize that these anomalous weights predominantly form the adversarial pathway.

\begin{figure*}
\centering  
\includegraphics[width=1\textwidth]{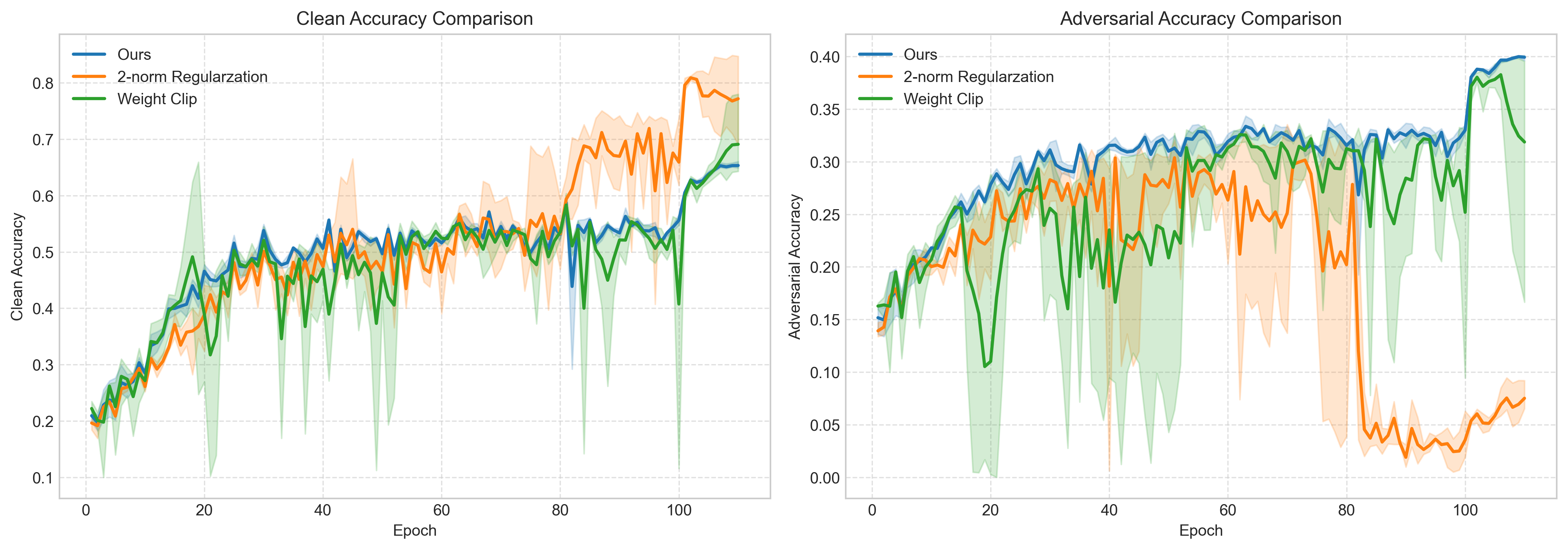} 
\caption{
\myadded{Ablation studies comparing our $\mathcal{L}_\text{reg}$ against simpler alternatives.}}
\label{fig10_clip}  
\end{figure*}

\myadded{We further conduct ablation studies to compare our method against simpler alternatives, including $\ell_2$ regularization and weight clipping based on the ratio between each weight and its mean (using the same threshold as our method). For each approach, we report results averaged over three independent runs in Fig. \ref{fig10_clip}. As can be observed, directly applying $\ell_2$ regularization fails to achieve stable optimization, where the model loses its adversarial robustness in the later stages of fast adversarial training. Weight clipping, while partially mitigating catastrophic overfitting, suffers from training instability and may relapse into catastrophic overfitting at any point. Moreover, robustness after clipping is slightly compromised. In contrast, our method consistently maintains stable performance throughout the training process.}

\section{Discussions and additional analyses}  
\label{DAA}

\begin{figure*}[htpb]
\centering  
\includegraphics[width=1\textwidth]{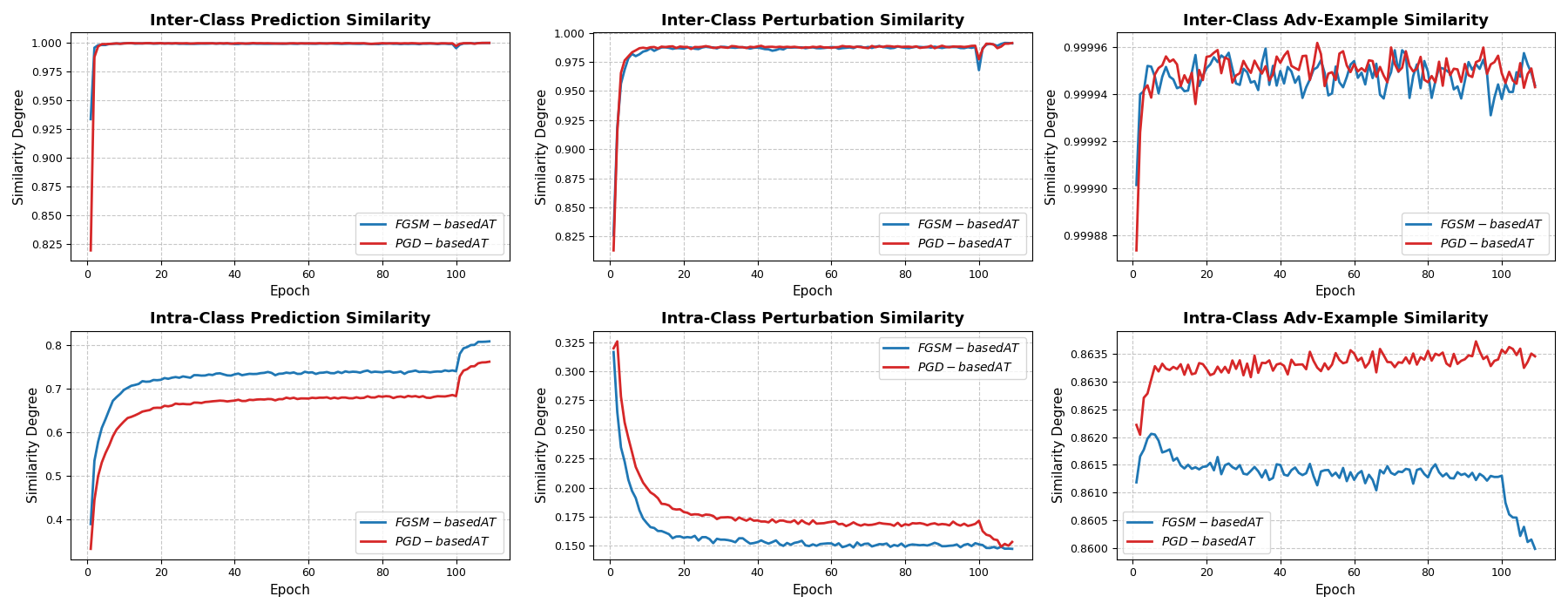} 
\caption{\myadded{Comparison between FGSM-based AT and PGD-based AT.}}  
\label{fig_similar}  
\end{figure*}

\textit{Why FAT improves robustness and why CO occurs.} 
\mydeleted{During the stable training phase in Figure 3, the adversarial loss $\mathcal{L}_\text{adv}$ exhibitshigher than the clean loss $\mathcal{L}_\text{clean}$. This indicates that adversarial
perturbations continue to challenge the model’s learned feature
representations, effectively suppressing the reliance on fragile
class-discriminative features. Consequently, the model is encouraged to distinguish categories with more robust features, leading to improved adversarial robustness. Notably, although
$\mathcal{L}_\text{adv}$ in FGSM-MEP shows a trajectory similar to $\mathcal{L}_\text{clean}$, it still
maintains a consistently higher value than $\mathcal{L}_\text{clean}$.}
\myadded{We further investigate why single-step FAT (e.g., FGSM-based AT) tends to induce trigger‑like overfitting, whereas multi-step AT (e.g. PGD-based AT \cite{xiong2020improved, zhong2024sparse, chen2021class}) does not. To this end, we conduct additional experiments comparing the two methods using multiple similarity metrics. As shown in Fig.~\ref{fig_similar}, we report inter-class and intra-class prediction similarity, adversarial perturbation similarity, and adversarial example similarity. Our results reveal three key observations. First, inter-class similarity exhibits comparable trends for both methods, indicating that they behave similarly across different learning stages. Second, intra-class perturbation similarity and intra-class adversarial example similarity are lower for FGSM-based AT than for PGD-based AT. This does not necessarily imply that the “trigger” patterns in FGSM perturbations are more consistent; rather, it may reflect that PGD, through its multi-step optimization, discovers more semantically coherent adversarial directions, leading to higher intra-class similarity in its perturbations. Third, intra-class prediction similarity is substantially higher for FGSM-based AT than for PGD-based AT. This suggests that although FGSM generates perturbations that appear more diverse (i.e., lower perturbation similarity), the model’s predictions on them are remarkably uniform. In contrast, for PGD-based AT, predictions on intra-class adversarial examples are less consistent, indicating that PGD introduces a wider variety of perturbation directions. Overall, the single-step optimization of FGSM restricts its adversarial directions, thereby causing CO.}

\textit{Transferability of methods across different tasks.}
Section \ref{sec:mitigation} shows that backdoor-inspired defense techniques can mitigate CO.
We further evaluate whether the method for CO can generalize to backdoor-related tasks.

For unlearnable techniques with sample-wise perturbations, adding uniform random noise to the poisoned dataset can effectively neutralize their impact. For example, training a ResNet-18 model on CIFAR with the original unlearnable dataset yields 13.58\% accuracy on clean samples, which increases to 94.22\% after applying random noise.
Hence, we focus on unlearnable techniques with class-wise perturbations. Prior work often addresses these using adversarial training \cite{fu2022robust}; for instance, a model trained on the original unlearnable dataset achieves 10.05\% accuracy on clean samples, whereas adversarial training improves accuracy to 86.89\%.
We investigate the transferability of the CO mitigation strategy to unlearnable tasks, with results summarized in Table~\ref{table14}. When the poisoning budget in unlearnable tasks is small, the $\mathcal{L}_{reg}$ for mitigating CO can also resolve the class-wise unlearnable attack, achieving performance comparable to the AT-based approach but at a significantly lower computational cost. However, under a large poisoning budget ($e.g.$ 8/255), its effectiveness diminishes.
These results further support viewing CO as a weak-trigger variant of unlearnable tasks.

\myadded{Additionally, the reduced effectiveness of $\mathcal{L}_{reg}$ under larger poisoning budgets (e.g., $8/255$) can be attributed to two key observations. First, the occurrence of weight outliers is not positively correlated with perturbation magnitude. In the extreme case where unconstrained adversarial examples are used, the unlearnable sample set is effectively transformed into a completely different dataset. Under this condition, the model exhibits no weight outliers, rendering $\mathcal{L}_{reg}$ completely ineffective. This suggests that the performance degradation of $\mathcal{L}_{reg}$ under larger perturbations arises because larger perturbations facilitate a direct association between the perturbation and the target label, thereby bypassing the need for weight anomalies. Second, our experiments focus on class-wise unlearnable perturbations, which tend to mimic the semantic features of a specific class. Once such perturbations successfully capture class semantics, the unlearnable samples again resemble a different dataset, resulting in the absence of weight outliers.}

\begin{figure*}
\centering  
\includegraphics[width=1\textwidth]{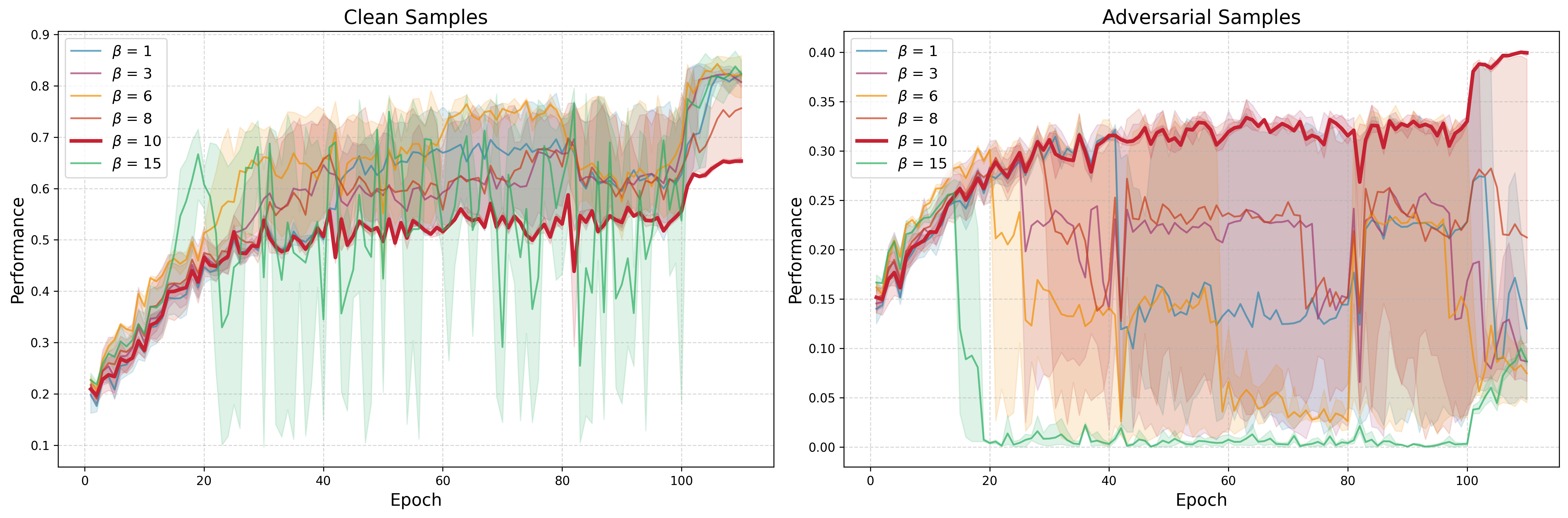} 
\caption{ \myadded{Ablation studies on the hyperparameter $\beta$.}}  
\label{fig_beta}  
\end{figure*}

\myadded{\textit{Impact of the hyperparameter $\beta$}.
The results in Figure~\ref{fig_beta} show that both too small and too large values of \(\beta\) adversely affect model robustness. Specifically, when \(\beta\) is too small, the regularization not only penalizes weight outliers but also constrains normal weights, thereby impairing the overall model performance. Conversely, when \(\beta\) is too large, the suppression of weight outliers becomes insufficient, rendering the regularization ineffective. }

\begin{table}[t]
\caption{Transferability of the CO mitigation strategy ($\mathcal{L}_{reg}$ in Eq.~(\ref{eq12}) ) to unlearnable tasks~\cite{huang2021unlearnable}. 
`Poisoned', `AT', and `$\mathcal{L}_{reg}$' correspond to models trained on the CIFAR-10 poisoned dataset using ResNet-18 with standard training, adversarial training, and standard training augmented with $\mathcal{L}_{reg}$, respectively. Each model is trained for 60 epochs using a cyclic learning rate schedule \cite{smith2017cyclical}. The reported values represent the final evaluation accuracy$\uparrow$ on clean samples.}
  \label{table14}
  \centering
  \setlength{\tabcolsep}{0.25cm} 
  \begin{tabular}{ c c c c }
    \toprule
Poisoning Budgets & Poisoned\cite{huang2021unlearnable} & AT\cite{fu2022robust}  & $\mathcal{L}_{reg}$ \\
\midrule
4/255 & 17.50 &  86.95  &  86.71    \\
6/255 & 10.05  &  86.89  &  86.48 \\
8/255 & 9.85  &  86.32 & 46.57 \\
\midrule
Training Time (minutes) & 15.4 & 36.8 & 15.5\\
    \bottomrule
  \end{tabular}
\end{table}

\section{Conclusions and future works}\label{conclusion}
In this work, we offer a systematic explanation of CO from a backdoor perspective, establishing a unified framework—trigger overfitting—that encompasses standard backdoor attacks, CO, and unlearnable tasks.
By validating phenomena such as pathway division, diverse path predictions, and the presence of universal class-distinguishable triggers in CO, we conceptualize CO as a weak-trigger variant of unlearnable tasks. Leveraging these insights to mitigate CO, we adapt several backdoor fine-tuning techniques to the FAT framework and propose a weight outlier suppression constraint. Experimental results validate our mechanistic explanation and demonstrate the effectiveness of backdoor-inspired strategies in alleviating CO.

Future research: This work has demonstrated that adversarial perturbations crafted by CO-affected models contain universal class-distinguishable (UCD) triggers. A natural issue arises: Can these UCD triggers be further extracted and explicitly identified? In other words, is it possible to deliberately induce the CO phenomenon with the extracted UCD triggers? Moreover, can triggers associated with other adversarial attacks be similarly extracted?
Models tend to overfit to triggers, resulting in high-confidence predictions on corresponding adversarial samples. Therefore, when defending against a specific attack, integrating manually extracted adversarial triggers can be more effective, as conventional adversarial training methods involve a trade-off between clean accuracy and robustness.

\nocite{langley00}

\bibliographystyle{IEEEtran}
\bibliography{ref}

\end{document}